%% file: main.tex
\documentclass[runningheads]{llncs}
\usepackage{graphicx}
\usepackage{booktabs}
\usepackage{subcaption}
\usepackage{amsmath}
\usepackage{multirow}
\usepackage{booktabs}
\usepackage{xcolor}
 \usepackage{enumitem}
\usepackage{soul}

\usepackage[ruled,linesnumbered]{algorithm2e}

\begin{document}
%
\title{\textsf{MORPH}: Towards Automated Concept Drift Adaptation for Malware Detection}
\titlerunning{\textsf{MORPH}}
%

\author{Md Tanvirul Alam\inst{1}, Romy Fieblinger\inst{1}, Ashim Mahara\inst{1}, and Nidhi Rastogi\inst{1}}
\authorrunning{Alam et al.}
%
\institute{Rochester Institute of Technology, Rochester NY 14623, USA\inst{1} \\
\email{\{ma8235, rf7344, am7539, nidhi.rastogi\}@rit.edu}
}

\maketitle              
\begin{abstract}
Concept drift is a significant challenge for malware detection, as the performance of trained machine learning models degrades over time, rendering them impractical.  While prior research in malware concept drift adaptation has primarily focused on active learning, which involves selecting representative samples to update the model, self-training has emerged as a promising approach to mitigate concept drift. Self-training involves retraining the model using pseudo labels to adapt to shifting data distributions. In this research, we propose MORPH -- an effective pseudo-label-based concept drift adaptation method specifically designed for neural networks. Through extensive experimental analysis of Android and Windows malware datasets, we demonstrate the efficacy of our approach in mitigating the impact of concept drift. Our method offers the advantage of reducing annotation efforts when combined with active learning. Furthermore, our method significantly improves over existing works in automated concept drift adaptation for malware detection. 

\end{abstract}

\input{sections/introduction}
\input{sections/motivation}

\input{sections/related-works}

\input{sections/methodology}

\input{sections/datasets}

\input{sections/experiments}
\input{sections/discussions}
\input{sections/conclusion}

%
%
\bibliographystyle{splncs04}
\bibliography{bibliography}

\end{document}

%% file: sections/introduction.tex
\section{Introduction}
Malware threats continue to evolve and pose a critical challenge to cyber defenses, requiring security experts to develop adaptable solutions for automated analysis and identification of malicious patterns. However, the effectiveness of these approaches depends on the statistical similarities between the trained defense models and the real-world data. Concept drift refers to a shift in the underlying distribution of the testing dataset that deviates from the training dataset (existing defense model). In the context of malware detection, this manifests in several ways, such as new malware variants emerging with malicious features or deliberate attempts by adversaries to evade existing detection methods~\cite{jordaney2017transcend}. The impact of concept drift can be far-reaching. Features that were earlier indicative of malware may become obsolete, leading to increased misclassification, false positives, or the overlooking of new threats~\cite{garcia2009anomaly,chen2020training}. It can deteriorate the effectiveness of defense models, rendering them impractical for further use~\cite{chen2023continuous}. It is crucial to note that concept drift is not a one-time event, and it requires continuous attention. Neglecting it may progressively impact defenses, creating a false sense of security just before a devastating cyberattack \cite{jordaney2017transcend,andresini2021insomnia}.

Prior research has addressed the onset of concept drift in various ways~\cite{gama2014survey}. While periodically retraining the models using the latest annotation dataset is an effective solution, obtaining sufficient high-quality annotated datasets for malware detection is often impractical~\cite{mohaisen2014av,wu2023grim}.
Due to these limitations, active learning has emerged as a widely adopted method for combating concept drift \cite{pendlebury2019tesseract}. The process of active learning selects a subset of representative samples from recent data, considering the challenges in labeling large volumes of data. The core idea is to choose samples the model is least confident about or those that ``deviate significantly from known data distribution"~\cite {jordaney2017transcend,barbero2022transcending}. Malware researchers then annotate these selected samples to ensure high-quality labels. As the model shows uncertainty towards these samples, they contribute significantly to adapting the decision boundary rapidly to the shifting data distribution. However, active learning methods still rely on the frequent availability of high-quality annotations, which can be costly to obtain~\cite{pendlebury2019tesseract}. To further alleviate the annotation effort, a model is retrained using ``pseudo labels"~\cite{lee2013pseudo,rizve2021defense} i.e., predictions made by the model itself. This approach, called weak supervision or self-training, although not extensively researched for malware detection, is emerging as a promising direction. We pursue these approaches in this paper.
\par
Notable prior works in automated concept drift adaptation for malware detection include DroidEvolver \cite{xu2019droidevolver} and its updated version, DroidEvolver++~\cite{kan2021investigating}, both designed specifically for Android malware classification. DroidEvolver\cite{xu2019droidevolver} maintains a pool of five classification models and utilizes the ensemble prediction as the pseudo-label to identify aging models that deviate from the ensemble. Subsequently, these models are retrained to align with the ensemble model. However, both these models face several challenges. Firstly, they rely on linear models, which can struggle with the high-dimensional, non-linear malware data distribution. Secondly, the performance of ensemble models degrades rapidly due to self-poisoning~\cite{kan2021investigating}, where incorrect pseudo-labels reinforce errors, leading to suboptimal adaptations.

\textbf{Main Contribution. }In this paper, we present \textbf{MORPH} (auto\ul{\textbf{M}}ated c\ul{\textbf{O}}ncept d\ul{\textbf{R}}ift ada\ul;{\textbf{P}}tation algorit\ul{\textbf{H}}m), a novel self-training neural network-based model for concept drift adaption in malware detection. \textit{Our approach leverages pseudo-labels to update the model, reducing the frequent need for active learning updates}. While pseudo-labels may introduce noise, they provide a significant advantage: the ability to utilize unlabeled data and adjust the model's decision boundaries accordingly~\cite{lee2013pseudo}. However, \textit{the inherent challenges of malware datasets require a specific pseudo-label sample selection strategy}, which we introduce in this research work. \textit{Our approach uses both labeled and pseudo-labeled data for semi-supervised learning within the neural network}. Furthermore, we investigate the effectiveness of combining pseudo-labels with active learning to enhance concept drift adaptation and improve malware detection performance. Specifically, we address the following \textbf{key research questions (RQ)} in this study:
\begin{enumerate}[leftmargin=*]


\item \textbf{RQ1:} \textit{Can pseudo-labeling alone enable automatic concept drift adaptation in neural network-based malware detection?}

\textit{To address this question}, we evaluate the effectiveness of pseudo-labels for capturing and adapting to evolving malware patterns without ground truth labels.


\item \textbf{RQ2:} \textit{To what extent can pseudo-label-based adaptation reduce the frequency of annotation in active learning?}

\textit{To explore this question}, we quantify the reduction in annotation requirements by comparing the performance of the pseudo-label-based adaptation algorithm with traditional active learning approaches.



\item \textbf{RQ3:} \textit{How does our pseudo-label-based approach with neural networks compare to prior automated concept drift adaptation methods?}

\textit{To investigate}, we benchmark our model performance against DroidEvolver~\cite{xu2019droidevolver} and DroidEvolver++~\cite{kan2021investigating} to demonstrate the efficacy of the proposed neural network-based method.

\end{enumerate}

We experimentally assess the efficacy of our approach on two benchmark malware detection datasets- an Android malware dataset~\cite{chen2023continuous} and a Windows malware dataset~\cite{2018arXiv180404637A}. Our rigorous experiments analyze the performance of our proposed approach on difference metrics, F1 score, False Positive Rate (FPR), and
False Negative Rate (FNR). Our approach and subsequent findings contribute insights into the efficacy of using pseudo-labels and active learning for effective concept drift adaptation in malware detection.

%% file: sections/motivation.tex
\section{Motivation}\label{sec-motivation}

\begin{figure}[t]
  \centering
  \begin{subfigure}[b]{0.4\textwidth}
    \includegraphics[width=\textwidth]{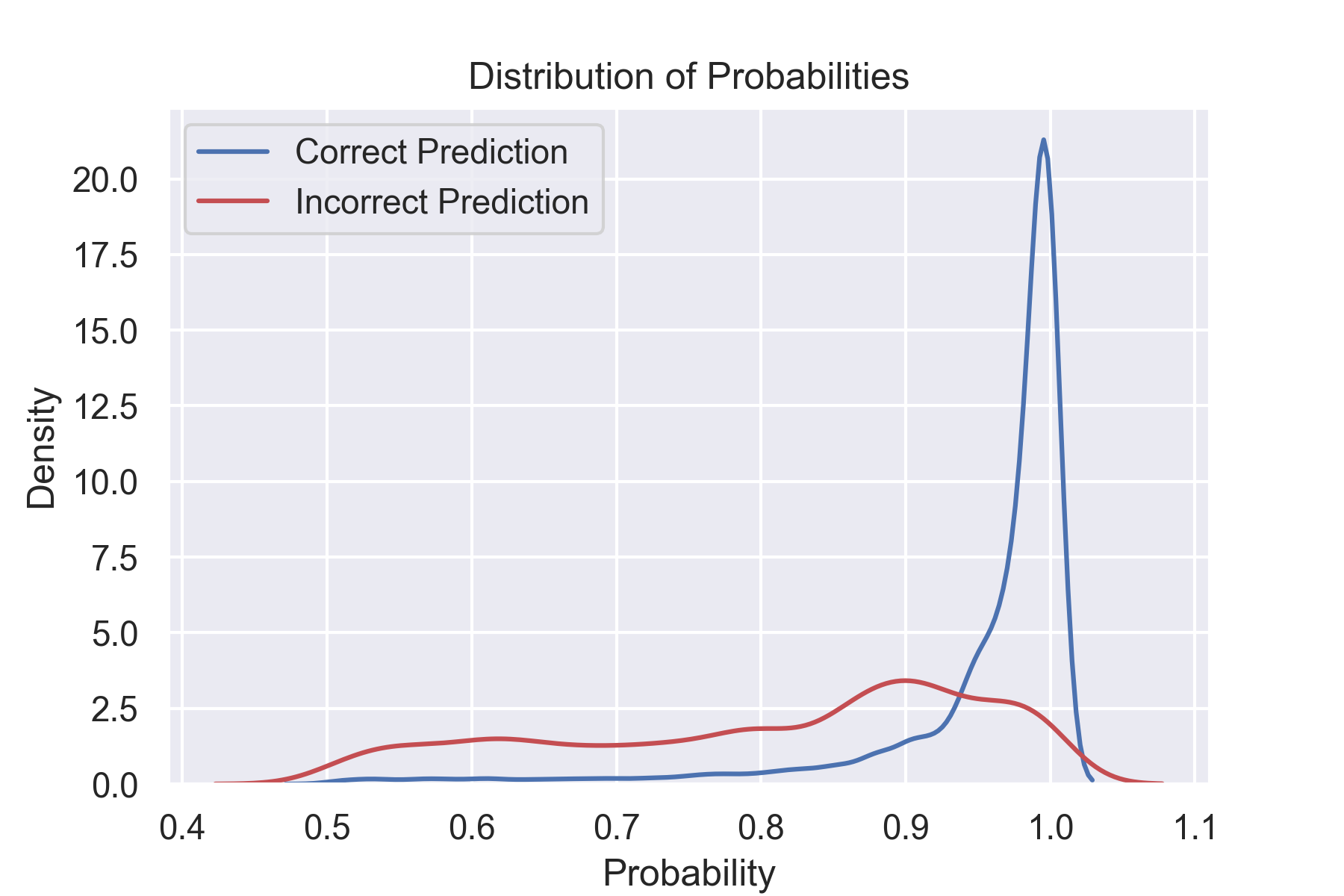}
  \end{subfigure}
  \begin{subfigure}[b]{0.4\textwidth}
    \includegraphics[width=\textwidth]{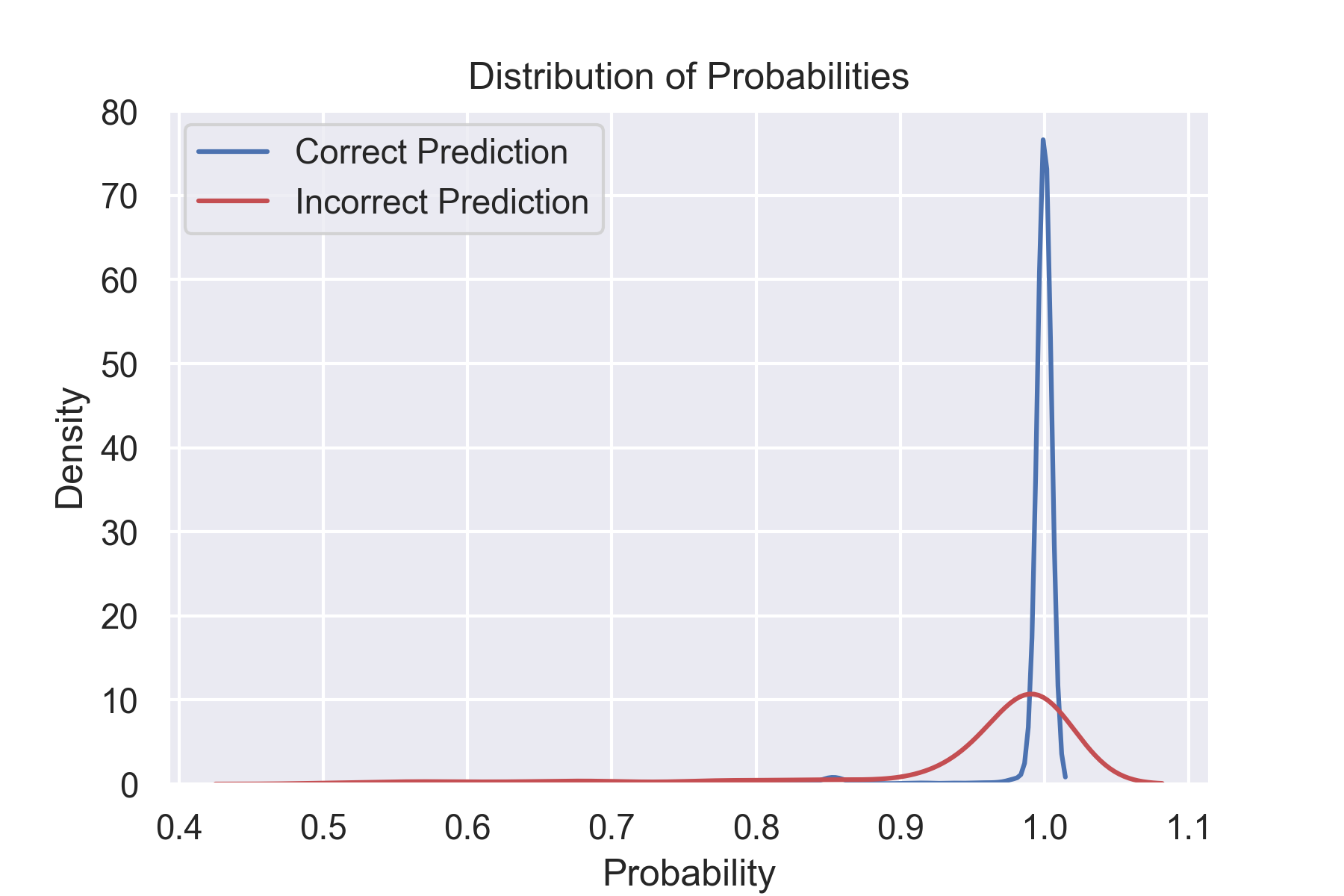}
  \end{subfigure}
  \caption{Kernel Density Estimate~\cite{chen2017tutorial} plot for probability distribution on Ember dataset~\cite{2018arXiv180404637A} for correctly \& incorrectly classified samples in Windows (left) and Android (right) application data.}
  \label{fig:confidence-correctness}
\end{figure}

Concept drift presents a significant challenge in malware detection, where the underlying data distribution is continuously evolving due to new malware families and malware variants being written frequently. Supervised learning-based malware detection models rely on malware application-labeled data for training, requiring extensive manual annotation efforts for both application and features (such as malicious behavior, code patterns, malware origin, and threat level)\cite{alam2023looking,rastogi2022tinker,alam2022cyner} to keep the detection model up-to-date. However, this approach is time-consuming and expensive since getting the latest malware files for training is not always practical. It is relatively easier to collect unlabeled application datasets than labeled ones for malware detection since manual annotation of malware requires significant effort from an expert~\cite{mohaisen2014av}. Therefore, pseudo-labeling can be an effective method in these situations as it leverages unlabeled data for semi-supervised learning~\cite{lee2013pseudo} when the labeled dataset available is small.
\par
Research shows that a neural network's predictions are more confident for correctly classified samples than incorrectly classified ones~\cite{staahl2020evaluation}. We illustrate this in Figure~\ref{fig:confidence-correctness}, which displays the softmax probability distribution plot for a malware classifier trained on a neural network using Android and Windows malware datasets. The correctly classified samples generally exhibit higher probabilities than the incorrect ones. However, there are instances where incorrect predictions can also have high confidence. This phenomenon arises from the lack of proper calibration in neural network predictions, which causes the output probabilities from a softmax layer to deviate from the true probabilities for model prediction~\cite{rizve2021defense}. So, even highly confident model predictions can sometimes be incorrect, leading to undetected malware.
Despite this calibration issue, neural networks often demonstrate some degree of robustness to label noise. Wu et al.~\cite{wu2023grim} have explored the application of semi-supervised learning to address the problem of `noisy labels' in malware detection. They modified a semi-supervised learning method designed initially for image classification~\cite{sohn2020fixmatch} and then demonstrated semi-supervised learning in mitigating the impact of noisy labels. However, further research on utilizing pseudo-labels to adapt neural networks-based malware detection models to concept drift is crucial.

\begin{figure}[t]
\centering
\includegraphics[width=0.8\textwidth]{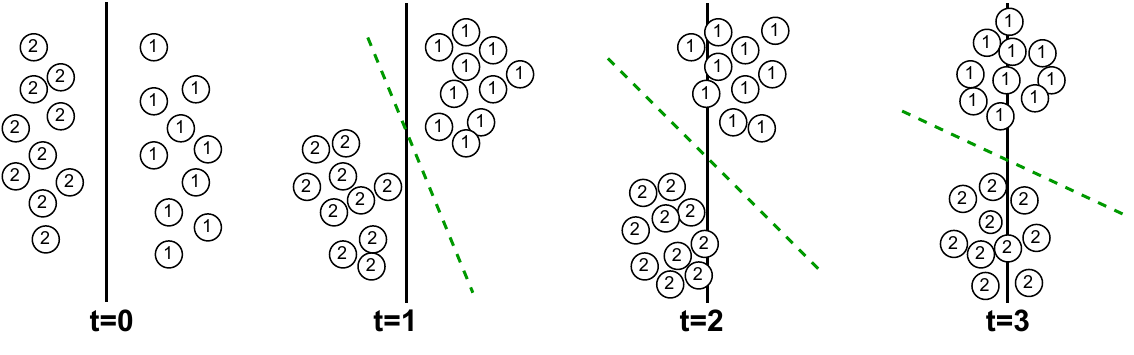}
\caption{Gradual adaptation to distribution shift for a binary classifier. The solid line represents the original classifier and the dotted lines represent the adapted classifier trained with pseudo-labels.}
\label{fig:gradual-drift}
\end{figure}

In a similar vein, self-training has been shown useful in gradual domain adaptation scenarios, especially where the data distribution undergoes gradual changes over time (such as malware data)~\cite{kumar2020understanding}. This phenomenon is effectively represented in Figure~\ref{fig:gradual-drift}. When we directly apply the source model to the target domain, as illustrated at time step t=3, its performance deteriorates, resulting in a random classifier due to the increasing mismatch between the source and target distributions. However, 
By gradually updating the model at intermediate steps, we can incorporate correctly classified pseudo labels into the training process, enabling the model to adjust its decision boundary in response to the evolving target domain. The iterative nature of self-training in concept drift allows the model to continuously refine its decision boundaries, enhancing its performance on the evolving target domain. By incorporating additional labeled samples obtained through active learning at regular intervals, the model can actively adapt to the changing data distribution. This motivates using active learning~\cite{chen2023continuous} with self-training for effectively addressing concept drift. Previous studies on adapting to malware concept drift have investigated active learning~\cite{chen2023continuous,barbero2022transcending}  or pseudo-labeling~\cite{xu2019droidevolver,kan2021investigating} separately. However, we are the first to explore the combination of these techniques in mitigating concept drift.

%% file: sections/related-works.tex
\section{Background and Related Work}
\label{sec:back}

\textbf{Concept Drift. }Concept drift refers to the phenomenon where the statistical properties of a target domain unpredictably shift~\cite{lu2014concept}, posing a critical challenge to machine learning models used in malware detection. Concept drift can manifest in multiple ways, including changes in feature distributions (e.g., new APIs)~\cite{zhang2020enhancing}, class distributions (e.g., new malware families)~\cite{zhang2020enhancing}, and even class definitions (e.g., evolving criteria for malicious behavior)~\cite{lu2014concept,kan2021investigating}. As the statistical properties of the target domain evolve, the model's performance can gradually deteriorate, rendering it unsuitable for accurate detection. In the malware domain, concept drift arises due to multiple reasons. When Android releases a major app update, it introduces new features and shifts distributions~\cite{android_version}. A significant culprit is the adversaries actively attempting to evade detection~\cite{aghakhani2020malware,anderson2018learning} by 
manipulating their software to trick detectors, leading to increased false negatives and a weakened defense system~\cite{aghakhani2020malware,anderson2018learning,pendlebury2019tesseract,kan2021investigating}.

\textbf{Pseudo-labeling. }Also called self-training, pseudo-labeling is a semi-supervised learning technique in which a security expert can assign malware labels to unlabeled application data based on malware detection model predictions. Mathematically, pseudo-labeling is defined as follows:

Let $\mathcal{D} = \{(\mathbf{x}_i, y_i)\}_{i=1}^{N_l}$ be the labeled dataset, where $\mathbf{x}_i$ represents the input sample and $y_i$ denotes its corresponding true label. Additionally, let $\mathcal{U} = \{\mathbf{x}_j\}_{j=1}^{N_u}$ be the unlabeled dataset, where $N_l$ and $N_u$ are the numbers of labeled and unlabeled samples, respectively.

Given a neural network model $f(\mathbf{x}; \theta)$ parameterized by $\theta$, the model's predictions for the unlabeled samples can be computed as $\hat{y}_j = f(\mathbf{x}_j; \theta)$.

Pseudo-labeling involves selecting a threshold $\tau$ to assign labels to the unlabeled samples based on the model's confidence in its predictions. The pseudo-label $\tilde{y}_j$ for an unlabeled sample $\mathbf{x}_j$ is defined as:

\[
\tilde{y}_j =
\begin{cases}
\arg\max_{k} \hat{y}_j^{(k)}, & \text{if } \max_{k} \hat{y}_j^{(k)} \geq \tau \\
\text{unlabeled}, & \text{otherwise}
\end{cases}
\]

where $\hat{y}_j^{(k)}$ denotes the predicted probability of the $k$-th class for the unlabeled sample $\mathbf{x}_j$. If the maximum predicted probability exceeds the threshold $\tau$, the sample is assigned the label corresponding to the class with the highest probability. Otherwise, the sample remains unlabeled. By incorporating these pseudo-labels into the training process, the semi-supervised learning algorithm can jointly optimize the model parameters $\theta$ using both the labeled and pseudo-labeled samples. 


\textbf{Semi-Supervised Learning (SSL). }It is halfway between supervised and unsupervised learning, where it utilizes both labeled data and unlabeled data for model training~\cite{chapelle2009semi}. The goal is to improve model accuracy beyond what is achievable with labeled data only by leveraging the unlabeled data to help the model distinguish the decision boundary between different classes. SSL works under several key assumptions~\cite{ouali2020overview}:
\begin{enumerate}
    \item If two data points in a high-density region are close, their corresponding output should also be close. Conversely, two points separated by a low-density region should have outputs that are distant from each other.
\item Data points belonging to the same cluster are likely to belong to the same class, implying that the decision boundary should lie in low-density regions.
\item The high-dimensional feature lies in a relatively low-dimensional manifold, enabling extraction of a low-dimensional representation through the use of unlabeled data to learn the simplified task.
\end{enumerate}

Prior work has used various SSL methods, such as consistency regularization, proxy labels, and generative models~\cite{van2020survey,yang2022survey}. 
Santos et al. used SSL to reduce the labeling effort required for malware detection from static features by 50\%~\cite{santos2011semi}. Mahdavifar et al.~\cite{mahdavifar2020dynamic} used semi-supervised learning with neural networks for malware category classification. MORSE, a deep neural network-based approach, leveraged SSL to learn from noisy labels of malware families~\cite{wu2023grim}. 

\textbf{Active Learning. }It aims to maximize the performance of a machine learning model by strategically selecting a minimal number of data samples for manual annotations, thereby minimizing labeling cost~\cite{ren2021survey}. It chooses samples that the model is most uncertain about~\cite{atlas1989training}. For instance, a binary classification task, like malware detection, might involve prioritizing data points with a posterior probability closest to 0.5 since they have the potential to improve the model's performance more than randomly selected data points. There are three major approaches for sample selection in active learning: uncertainty-based, diversity-based, and expected model change. The uncertainty-based approach defines and measures uncertainties of new data points, selecting the least confident ones for annotation~\cite{joshi2009multi,tong2001support}. The diversity-based approach focuses on selecting data points that better represent the overall distribution of unlabeled samples~\cite{nguyen2004active,sener2017active}. Lastly, the expected model change approach selects samples that would have the most significant impact on the current model parameters~\cite{roy2001toward}.
\par
While active learning is relatively understudied in the context of malware detection, uncertainty sampling has emerged as the prevalent popular approach. Equivalently, this can be viewed as a form of out-of-distribution (OOD) sample selection, targeting data samples with the highest uncertainty scores. Prediction confidence is a commonly used baseline for OOD sample selection. However, since the model's predicted probabilities may not accurately reflect its confidence, the predictions require calibration~\cite{guo2017calibration}.
Transcend~\cite{jordaney2017transcend}, a parametric statistical framework for assessing model uncertainty and identifying potential concept drift through nonconformity measures, leverages two metrics, namely model confidence, and credibility, to reject test samples that may have experienced drift. Transcendent~\cite{barbero2022transcending} improves upon Transcend by refining the framework and introducing two new conformal evaluators, significantly reducing computational overhead. However, both Transcend~\cite{jordaney2017transcend} and Transcendent~\cite{barbero2022transcending} assume that data distributions remain stationary within certain time windows. This assumption might not always hold in real-world scenarios where drift can occur gradually or in more complex patterns.

%% file: sections/methodology.tex
\section{Proposed Methodology}
\label{sec:met}

\begin{figure}[t]
\centering
\includegraphics[width=0.8\textwidth]{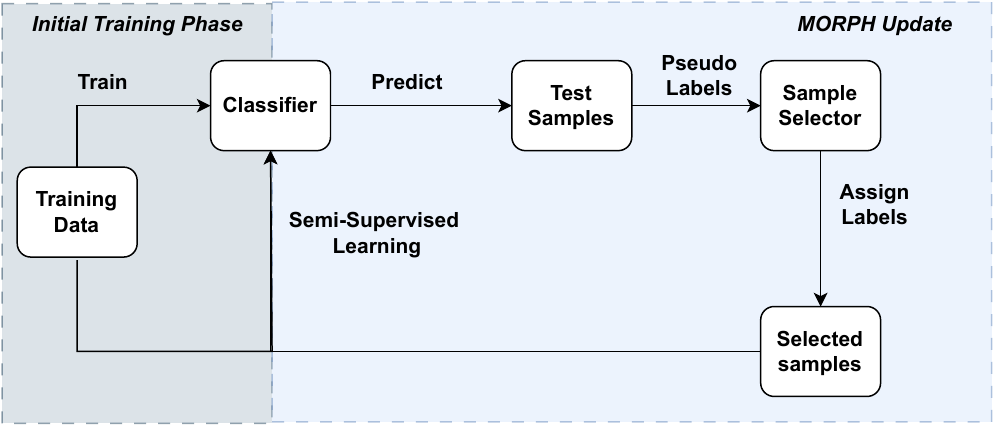}
\caption{Proposed concept drift adaptation algorithm, MORPH}
\label{fig:method}
\end{figure}


Figure~\ref{fig:method} presents MORPH, our methodology for adapting to concept drift in malware detection. Our approach deploys a continual learning neural network model, initially trained on labeled samples. Since it is crucial to retrain at regular intervals, and while existing work has explored optimal timing, our work focuses on a simplified scenario for simplicity. Following Chen et al.~\cite{chen2023continuous}, we retrain the model monthly using new samples, generating pseudo-labels from the model predictions on unlabeled data. Our novel sample-selection algorithm then chooses malware and benign samples for inclusion in the training process, and the model is retrained via semi-supervised learning, leveraging both original labeled and pseudo-labeled data.


\subsection{Sample Selection Algorithm}
\label{sec:sample-selection}

\begin{figure}[t]
  \centering
  \begin{subfigure}[b]{0.4\textwidth}
    \includegraphics[width=\textwidth]{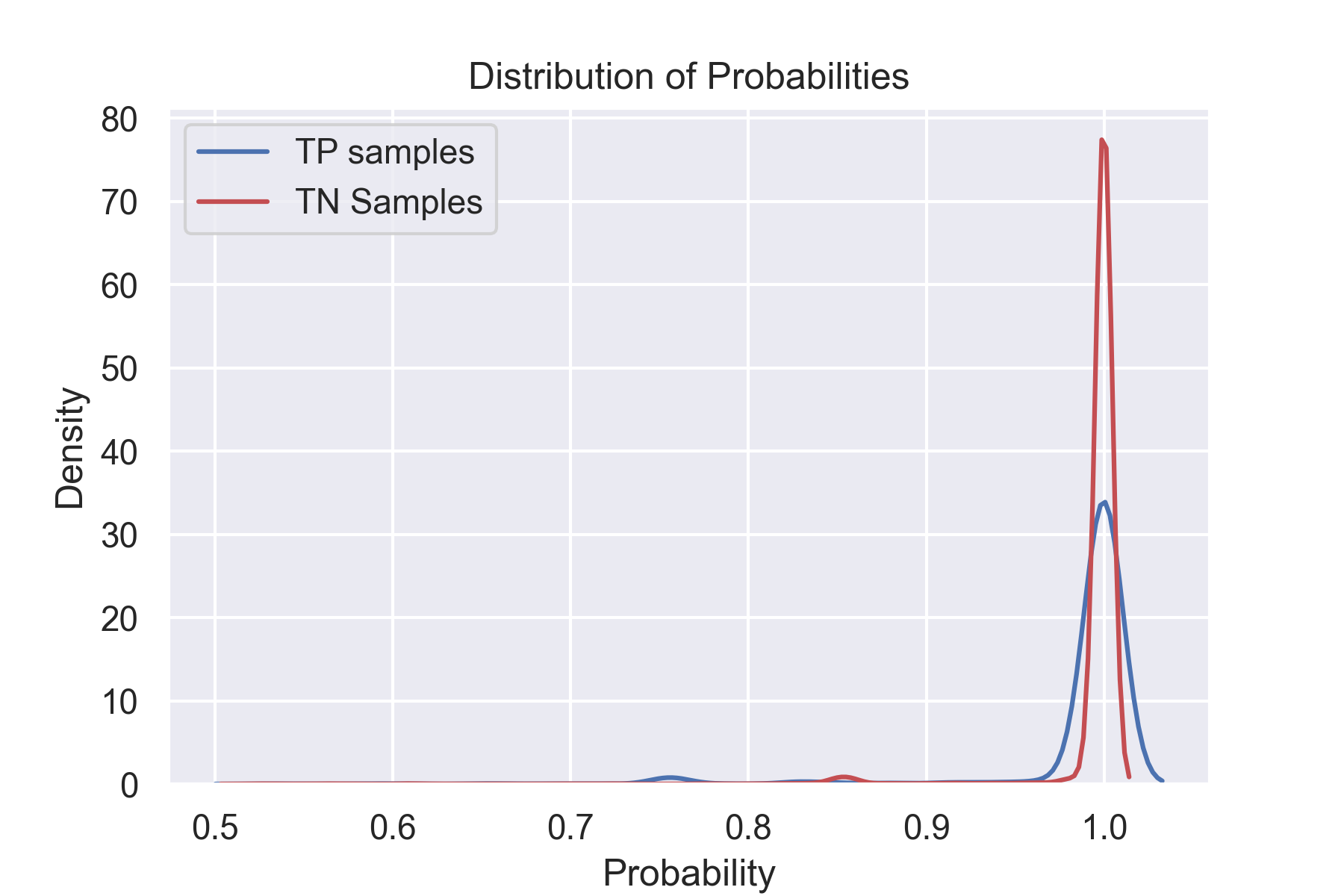}
  \end{subfigure}
  \begin{subfigure}[b]{0.4\textwidth}
    \includegraphics[width=\textwidth]{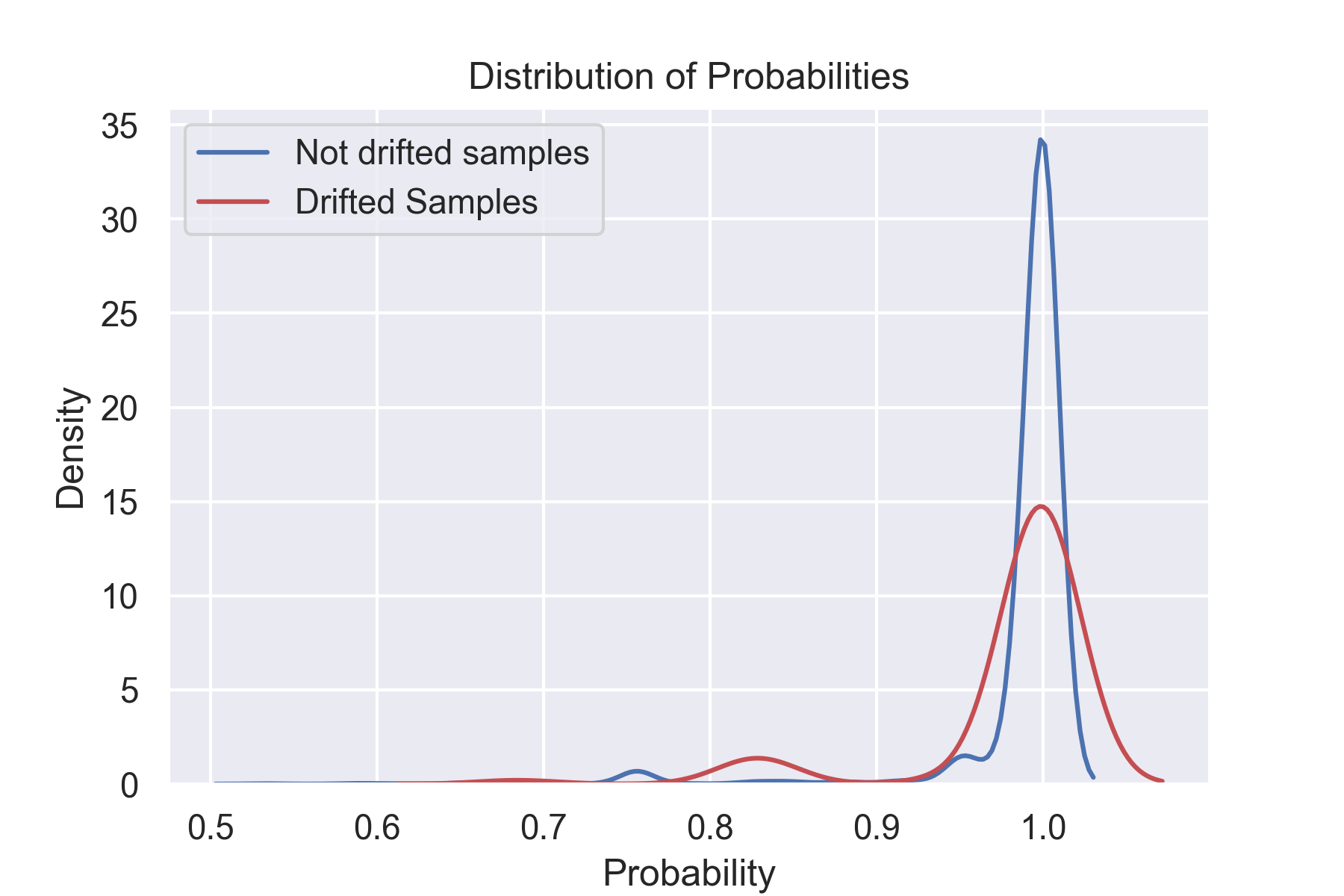}
  \end{subfigure}
  \caption{Kernel Density Estimate~\cite{chen2017tutorial} plot for probability distribution on AndroZoo~\cite{Allix:2016:ACM:2901739.2903508} dataset for (left) True Positive and True Negative sample and (right) drifted vs not-drifted malware samples}
  \label{fig:motivation2}
\end{figure}

The unique threat of concept drift in malware classification is its bias towards false negatives, where malware instances are erroneously classified as benign. This is a consequence of, for example, malware authors actively attempting to evade existing detection methods, resulting in significant variations between newer and older malware characteristics. As a result, when a model predicts a new sample as malware, the likelihood of being correct is higher due to the relatively low occurrence of false positives. Therefore, even predictions with low confidence tend to be accurate.
Conversely, low-confidence predictions are more likely to be incorrect in the case of benign predictions. This phenomenon is illustrated on the left half of Figure~\ref{fig:motivation2}, where the malware predictions (true positives) have a broader probability distribution compared to benign predictions (true negatives). This suggests that low-confidence malware classifications are likely accurate due to the rarity of false positives. Similarly, the right side of Figure~\ref{fig:motivation2} further reveals that drifted malware samples, which deviate from the training data, exhibit even broader distributions. These observations emphasize the importance of treating benign and malware samples and their associated pseudo-labels differently. 

To build on this insight, we propose a targeted pseudo-label strategy. For predicted malware, we prioritize incorporating uncertain instances into training. This exposure allows the model to adapt to emerging threats and evolving malware characteristics. Conversely, for predicted benign samples, we prioritize high-confidence predictions to minimize erroneous inclusions. Through this asymmetry, we effectively address the unique complexities of concept drift in malware detection.


\begin{algorithm}[t]
\SetAlgoLined
\SetKwInOut{Input}{Input}
\SetKwInOut{Output}{Output}
\Input{Test samples}
\Output{Pseudo-labeled malware and benign samples}
\BlankLine
Calculate model's prediction on test samples\

$D_M \leftarrow \text{samples predicted as malware}$\

$D_B \leftarrow \text{samples predicted as benign}$\
\BlankLine
Select $N_M$ samples randomly from $D_M$ with probability $> \tau_m$\
\BlankLine
Select top $N_M$ samples with highest confidence from $D_B$ with probability $> \tau_b$\
\BlankLine
Return the pseudo-labeled malware and benign samples selected\
\caption{MORPH: Pseudo-labeling Algorithm}
\label{algo:pseudo-labeling}
\end{algorithm}
\par
Algorithm~\ref{algo:pseudo-labeling} outlines the steps involved in the sample selection algorithm. In \textit{Step 1}, the algorithm predicts the model's labels for the unlabeled test data. Based on these predicted probabilities, pseudo-labeled samples are generated. In \textit{Step 2}, a selected number, $N_M$, of samples is chosen from the predicted malware samples, using a probability threshold, $\tau_m$. The value of $N_M$ can be adjusted as a hyperparameter to control the utilization of unlabeled data, especially in the case of large datasets. Similarly, in \textit{Step 3}, an equal number of samples are selected from the predicted benign samples, but with a different confidence threshold, $\tau_b$, where $\tau_m$ is smaller than $\tau_b$.
In \textit{Step 4}, we randomly sample pseudo-labeled malware samples as long as they are above $\tau_m$, ensuring drifted samples are included. However, in Step 5, for benign samples, the algorithm prioritizes the selection of the most confidently predicted instances. Finally, in \textit{Step 6}, the hyperparameter $\tau_b$ may be omitted when a significantly higher number of samples are predicted as benign compared to malware. This is often the case in malware datasets due to the larger proportion of benign applications in the wild~\cite{jordaney2017transcend}. By sampling an equal number of malware and benign samples, the value of $\tau_b$ approaches 1, effectively diminishing the inclusion of false negative samples during the training phase.

\subsection{Semi-Supervised Training}
After the sample detector identifies relevant instances, we leverage semi-supervised learning to retrain the model by combining both ground truth and pseudo-labeled samples. To achieve this, we form mini-batches comprising an equal number of original ground truth data and pseudo-labeled data. During training, the objective function balances two losses:
\begin{equation*}
    L = L_s + \lambda L_u
\end{equation*}
Here, $L_s$ represents the standard supervised loss for the manually annotated samples, while $L_u$ represents the loss for the pseudo-labeled samples. The hyperparameter $\lambda$ controls the relative contribution of each loss. During the training phase, we select an equal number of samples from the labeled and unlabeled data, ensuring that one sample from each dataset is seen during one epoch of training on the labeled data. Based on empirical observation, we set $\lambda$ to 1.

%% file: sections/datasets.tex
\section{Datasets} 
\label{sec:data}

\begin{table}[t]
\centering
\begin{tabular}{|c|ccc|ccc|ccc|}
\hline
\multirow{2}{*}{\textbf{Dataset}} & \multicolumn{3}{c|}{\# Months} & \multicolumn{3}{c|}{Benign Apps} & \multicolumn{3}{c|}{Malicious Apps} \\ \cline{2-10} 
 & \multicolumn{1}{c|}{Train} & \multicolumn{1}{c|}{Dev} & Test & \multicolumn{1}{c|}{Train} & \multicolumn{1}{c|}{Dev} & Test & \multicolumn{1}{c|}{Train} & \multicolumn{1}{c|}{Dev} & Test \\ \hline
\textbf{AndroZoo} & \multicolumn{1}{c|}{12} & \multicolumn{1}{c|}{6} & 18 & \multicolumn{1}{c|}{40947} & \multicolumn{1}{c|}{18109} & 30797 & \multicolumn{1}{c|}{4542} & \multicolumn{1}{c|}{2028} & 3631 \\ \hline
\textbf{Ember} & \multicolumn{1}{c|}{1} & \multicolumn{1}{c|}{1} & 22 & \multicolumn{1}{c|}{17180} & \multicolumn{1}{c|}{32820} & 650000 & \multicolumn{1}{c|}{32761} & \multicolumn{1}{c|}{27239} & 730904 \\ \hline
\end{tabular}
\caption{Summary of AndroZoo and Ember datasets}
\label{tab:dataset}
\end{table}

We employ two distinct datasets-- AndroZoo~\cite{Allix:2016:ACM:2901739.2903508} and EMBER~\cite{2018arXiv180404637A}, to experimentally evaluate our approach, MORPH, for concept drift in malware detection on Android and Windows platforms. The AndroZoo dataset is a comprehensive collection of Android applications, including millions of APKs analyzed by multiple AntiVirus products to identify malware. On the other hand, EMBER is a labeled benchmark dataset used for training machine learning models to statically detect malicious Windows portable executable files, with features extracted from over a million binary files. The summaries of these two datasets are presented in Table~\ref{tab:dataset} and further details are provided below. 

\subsection{Android dataset}This dataset was introduced by Chen et al.~\cite{chen2023continuous} and is derived from AndroZoo~\cite{Allix:2016:ACM:2901739.2903508}. It consists of Android applications spanning three years from 2019 to 2021 and consists of 16,978 binary features indicating different API usage. Each application in the dataset has over 20 types of metadata, such as VirusTotal reports. We use the data from the first year to construct our training dataset, while the data from the first month of 2018 serves as the validation dataset. The data from the remaining 23 months, spanning the second and third years, is set aside for the testing. The training dataset comprises 121 malware families, whereas the test datasets from 2020 and 2021 include 82 and 51 families, respectively.

\subsection{Windows Dataset} We use the EMBER dataset for Windows PE malware detection~\cite{2018arXiv180404637A}. This dataset offers features extracted from each file using the LIEF project~\cite{LIEF}. Each file is characterized by a 2381-dimensional feature vector that encapsulates various aspects of the executable file: byte-level sequences, imported functions, and header information. We make use of the features from the years 2017 and 2018. We train our model using data from the first month of 2017, while data from the second month of 2017 is used for validation. The remaining 22 months of data are used for testing. The 2018 samples in the dataset were deliberately selected for their classification difficulty, leading to the expectation of bad performance from the model trained on earlier data during the test month. During our experiments, we standardize the feature set using the Scikit-learn StandardScaler.

%% file: sections/experiments.tex
\section{Experiments and Results}

\subsection{Experimental Settings}
In our experimental setup, we conducted training and validation using two distinct datasets: AndroZoo~\cite{Allix:2016:ACM:2901739.2903508} for Android applications and Ember~\cite{2018arXiv180404637A} for Windows applications. For the AndroZoo dataset, we trained our model using the first 12 months of available data. However, due to the larger size of the Ember dataset and its nearly equal proportion of malware and benign samples, we trained the model using only the first month of training data from Ember~\cite{2018arXiv180404637A}.

Our baseline classifier is a multilayer perceptron that takes input features of dimension 2381 for the EMBER and 16,978 for the AndroZoo. The hidden layer dimensions are 512-384-256-128. We fine-tuned the batch size, learning rate, dropout rate, and training epochs separately for each dataset using their respective validation sets to optimize the model's performance.

Following the training phase, we evaluated the model's performance on the remaining test data. We measured the F1 score, False Positive Rate (FPR), and False Negative Rate (FNR) for each test month. Instead of training from scratch each month, we performed fine-tuning to adapt the model to concept drift. Importantly, we retained the original training data during each test month. This approach served two purposes: firstly, it helped mitigate the potential impact of self-poisoning \cite{kan2021investigating} by using high-quality ground truth annotations for the selected samples, and secondly, it prevented catastrophic forgetting that may occur when removing the knowledge of the initial malware samples \cite{rahman2022limitations}.

We identified the optimal value for $\tau_m$ using the AndroZoo dataset since we noticed a more severe drift in the dataset, resulting in poorer model performance. We use the first six months of data for this, following the approach in~\cite{chen2023continuous}, which used the same dataset for optimizing parameters for the active learning algorithm. We use the obtained $\tau{_m}=0.6$ for AndroZoo and EMBER to prevent overfitting on a separate dataset. As mentioned in Section~\ref{sec:sample-selection}, we can omit the parameter $\tau_b$ as more samples are predicted as benign due to a more significant proportion of benign samples and the presence of more false negatives in the test dataset.

\subsection{Utility of Pseudo Labels}

\begin{figure}[t]
  \centering
  \begin{subfigure}[b]{0.32\textwidth}
    \includegraphics[width=\textwidth]{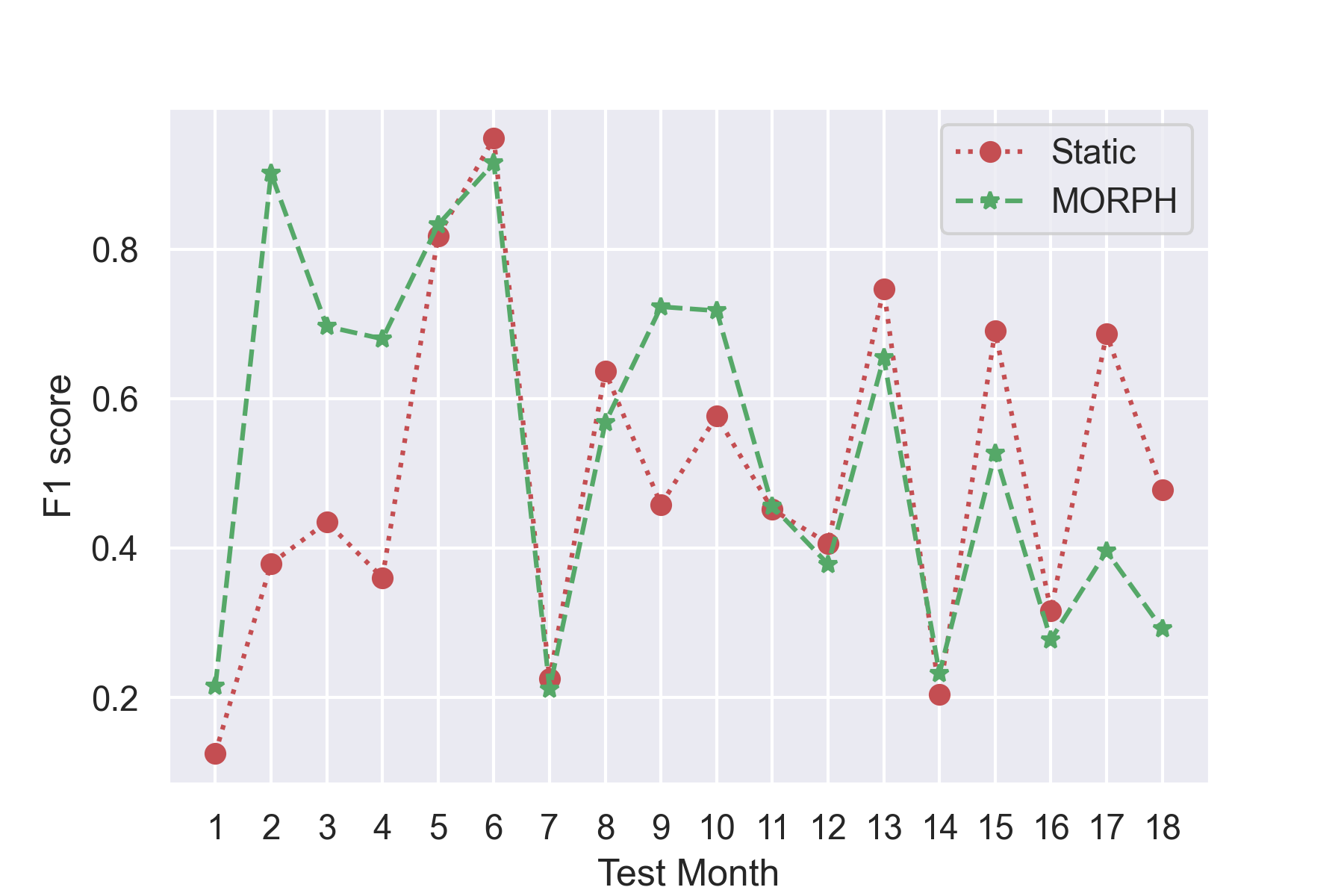}
  \end{subfigure}
  \hfill
  \begin{subfigure}[b]{0.32\textwidth}
    \includegraphics[width=\textwidth]{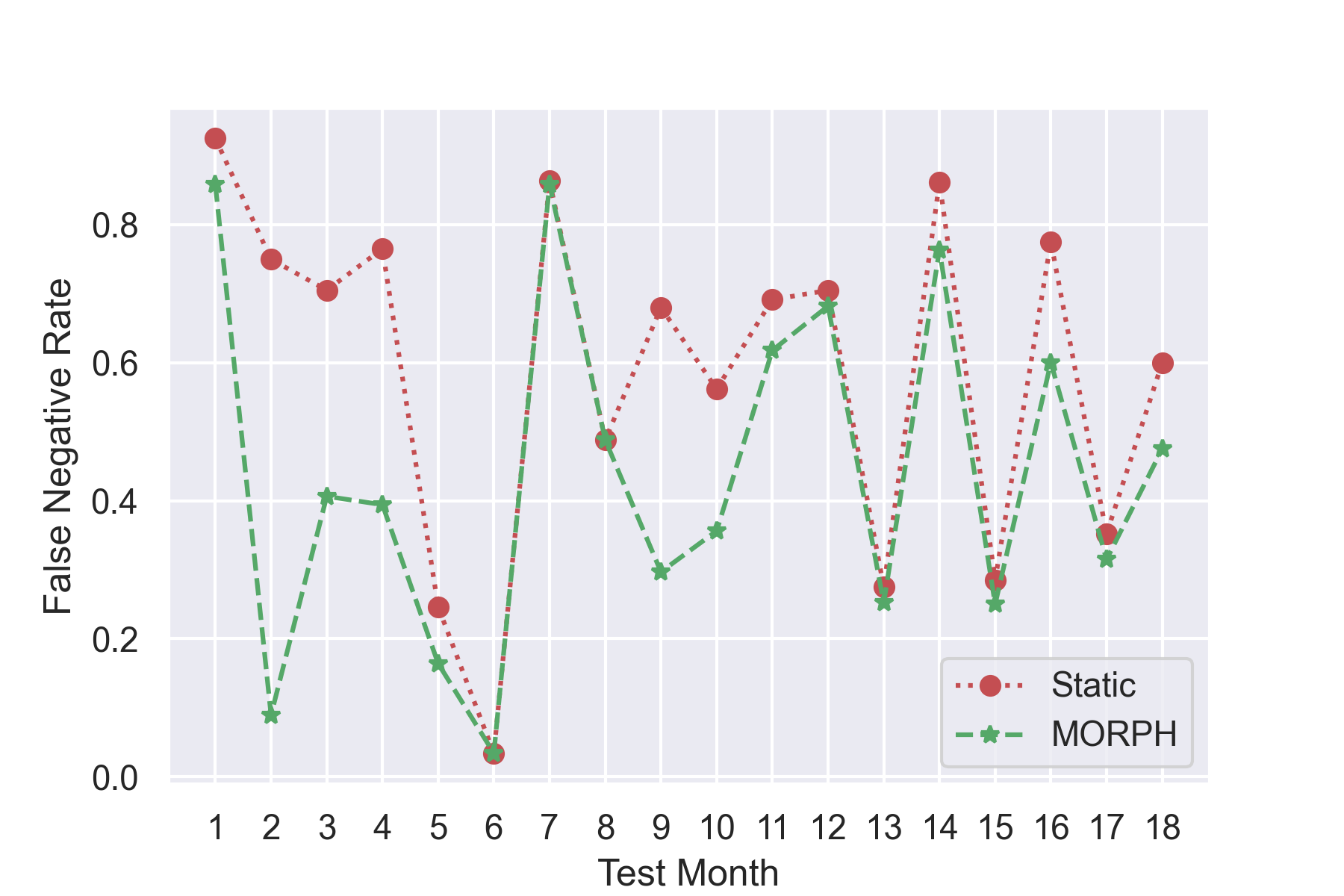}
  \end{subfigure}
    \hfill
  \begin{subfigure}[b]{0.32\textwidth}
    \includegraphics[width=\textwidth]{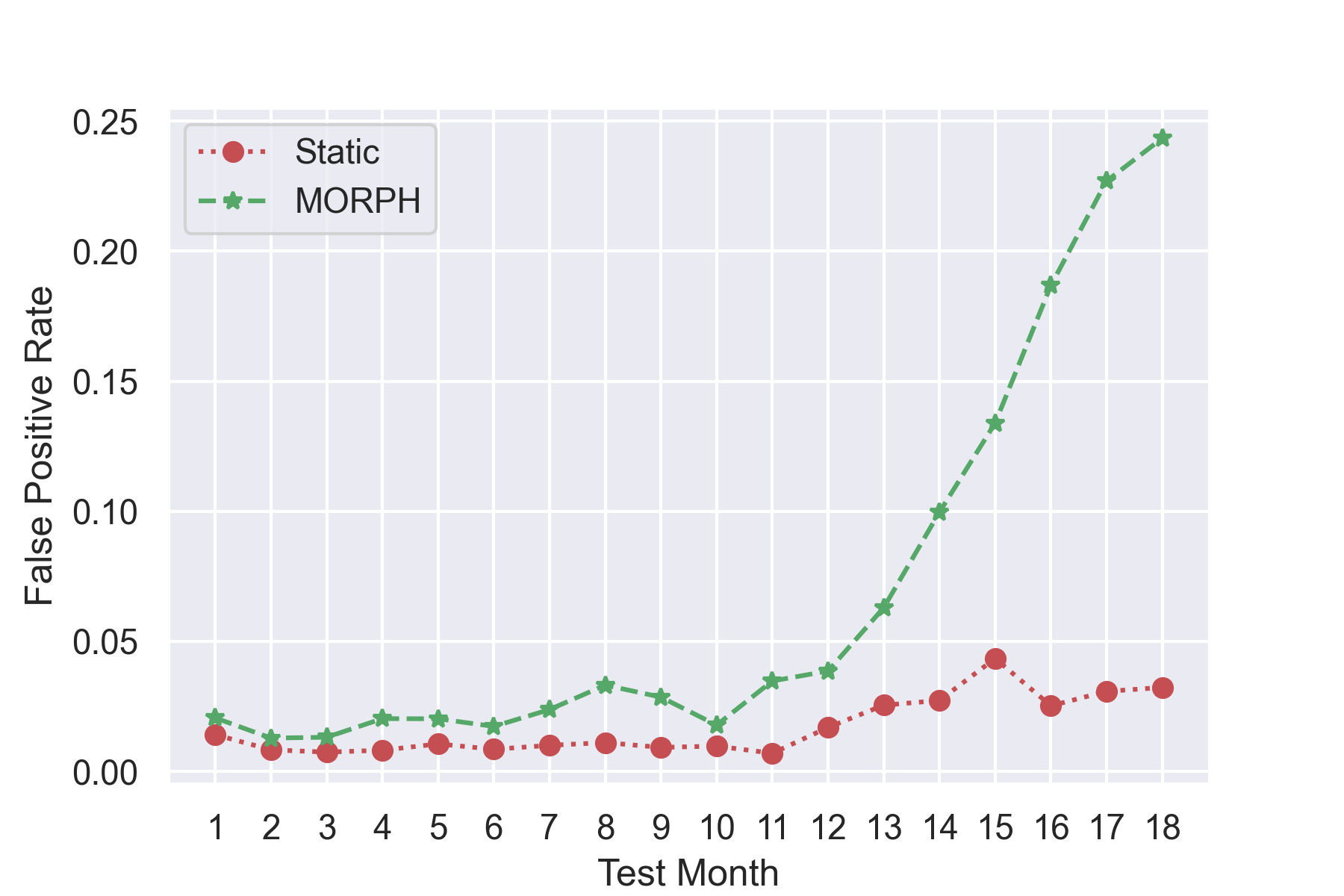}
  \end{subfigure}
  \caption{F1 score (left), FNR (middle), and FPR (right) for test months on AndroZoo dataset with MORPH and baseline (Static) neural network.}
  \label{fig:androzoo-res}
\end{figure}

\begin{figure}[t]
  \centering
  \begin{subfigure}[b]{0.32\textwidth}
    \includegraphics[width=\textwidth]{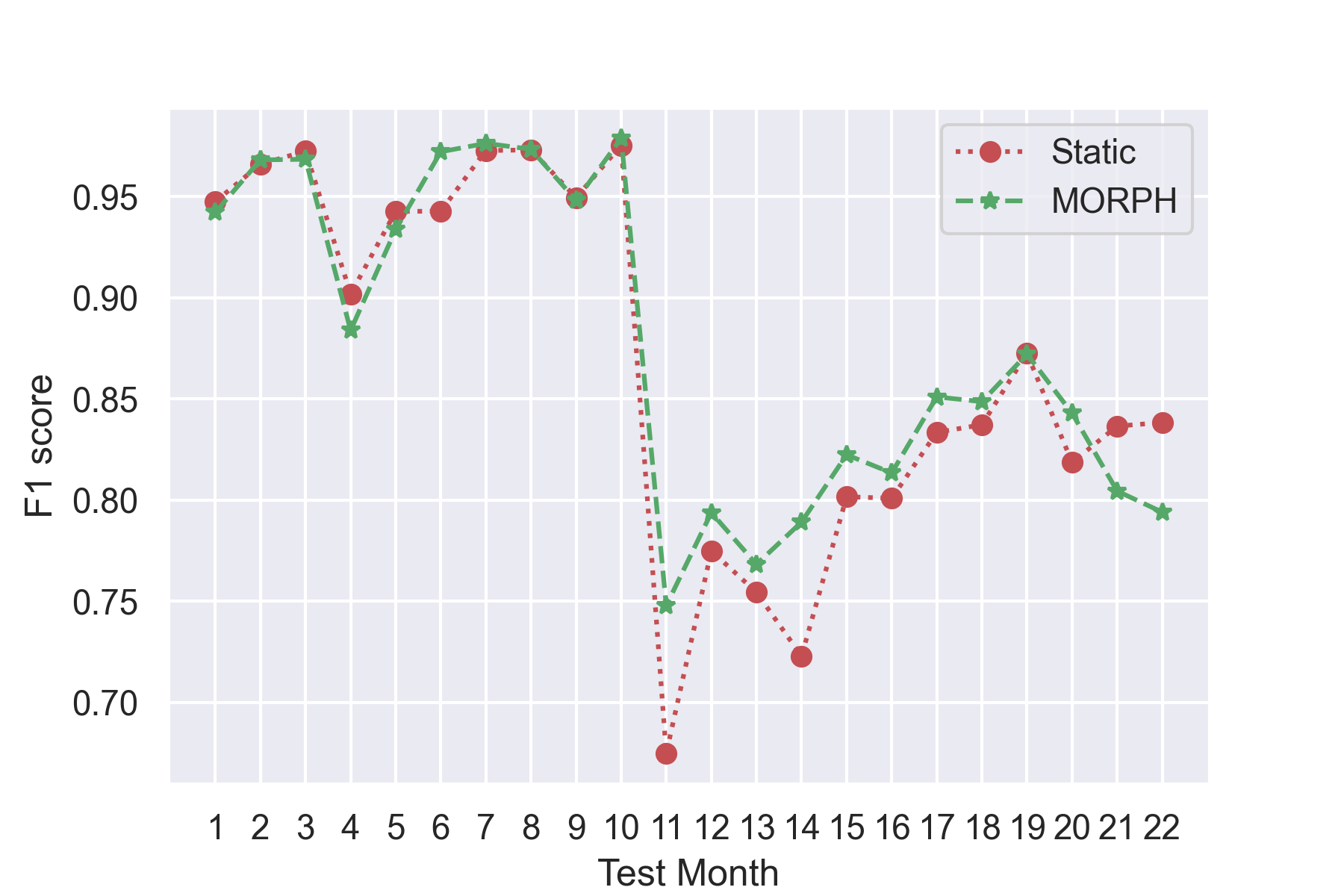}
  \end{subfigure}
  \hfill
  \begin{subfigure}[b]{0.32\textwidth}
    \includegraphics[width=\textwidth]{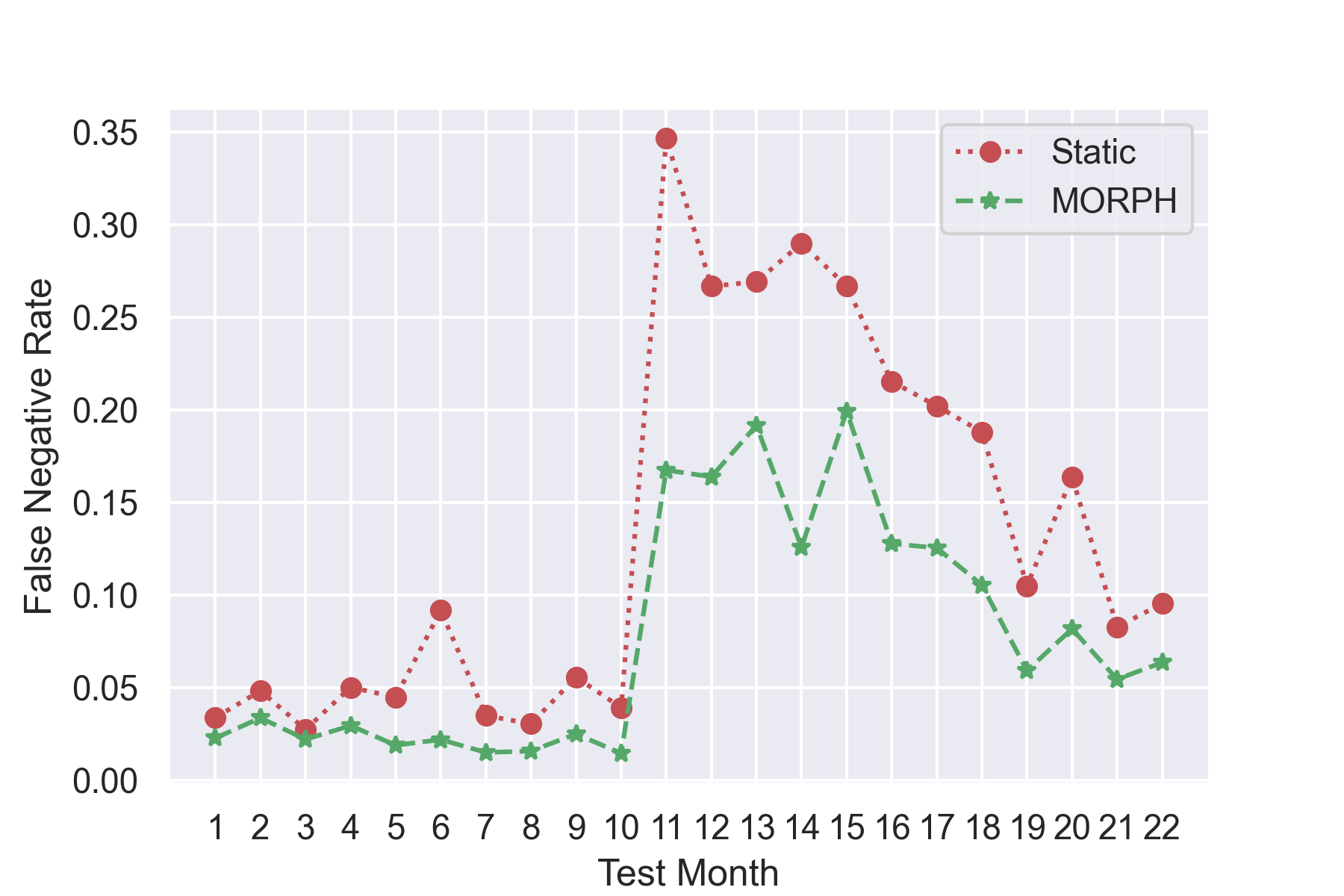}
  \end{subfigure}
    \hfill
  \begin{subfigure}[b]{0.32\textwidth}
    \includegraphics[width=\textwidth]{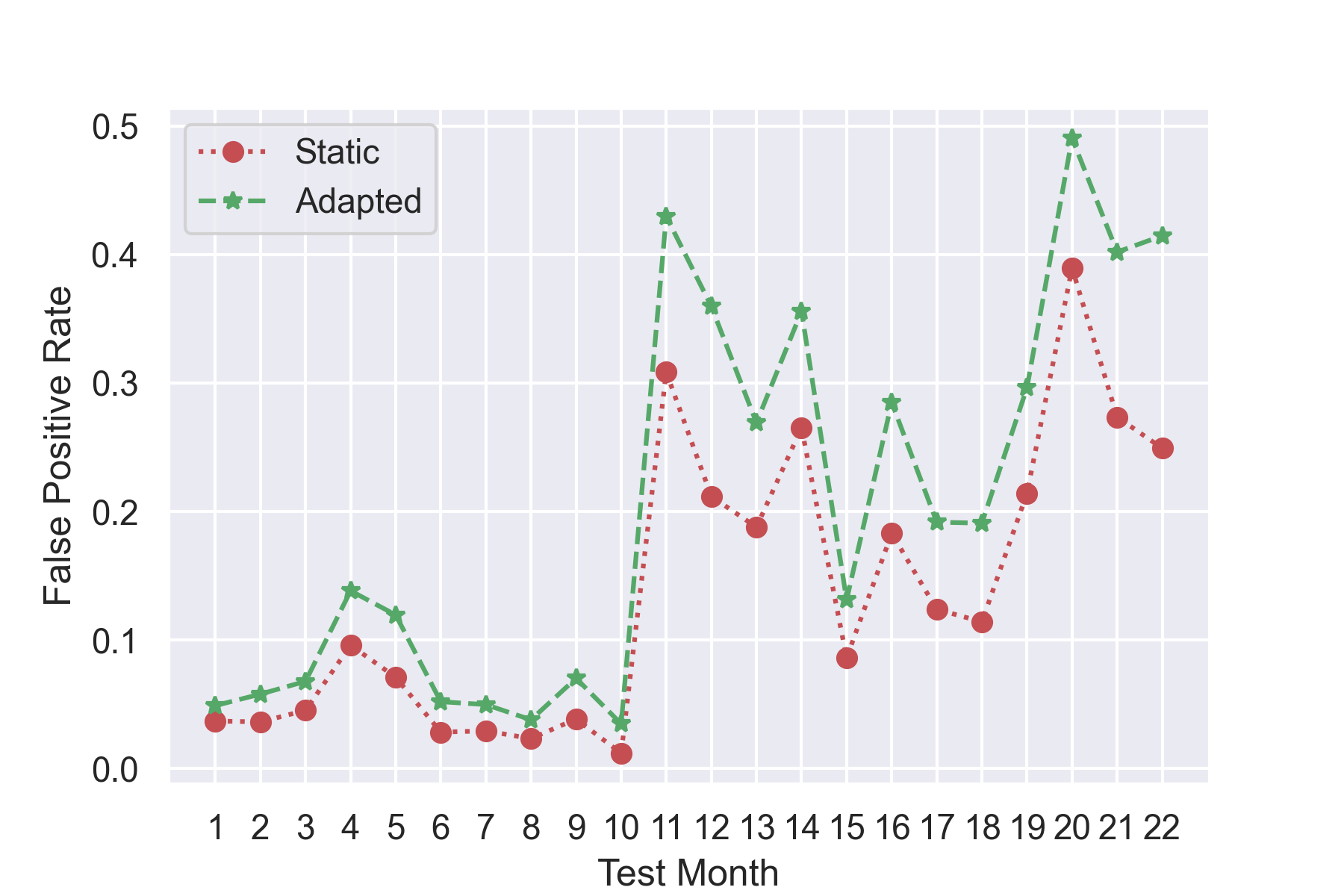}
  \end{subfigure}
  \caption{F1 score (left), FNR (middle), and FPR (right) for test months on Ember dataset with MORPH and baseline (Static) neural network.}
  \label{fig:ember-res}
\end{figure}

The performance of our proposed method is shown in Figure~\ref{fig:ember-res} and Figure~\ref{fig:androzoo-res} for the Ember and AndroZoo datasets, respectively. For the Ember dataset~\cite{2018arXiv180404637A}, we observe a significant performance drop in test month 11, which corresponds to data from January 2018. This decline was expected, as the data collected in 2018 presented more significant challenges for the classification model, as noted in~\cite{2018arXiv180404637A}. However, our proposed method effectively recovers performance in the subsequent months, as evidenced by improved F1-score and lower false negative rate (FNR). Similarly, for the AndroZoo dataset, our adaptation method outperforms the baseline neural network following significant drift. We can effectively incorporate drifted samples into the model by leveraging semi-supervised learning. While we observe a slight self-poisoning effect towards the end, our proposed method demonstrates greater robustness than prior work, as discussed in Section~\ref{sec:de-cmp}, as the onset of such poisoning effects is significantly delayed.


Overall, our proposed method yields substantial improvements over the baseline neural network. On average, we achieve a 4.07\% increase in F1-score for the AndroZoo dataset and a 0.83\% increase for the Ember dataset. Moreover, the false negative rate is reduced by 14.78\% and 5.74\% for the AndroZoo and Ember~\cite{2018arXiv180404637A} datasets, respectively.

\textbf{\textit{We arrive at the following conclusions for our RQ1}}

\begin{enumerate}
    \item Pseudo labels generated by neural networks can provide sufficient information for concept drift adaptation for malware detection.
    \item As the drift becomes more severe, pseudo labels can provide more benefits to improve model performance.
\end{enumerate}


\begin{figure}[t]
  \centering
  \begin{subfigure}[b]{0.32\textwidth}
    \includegraphics[width=\textwidth]{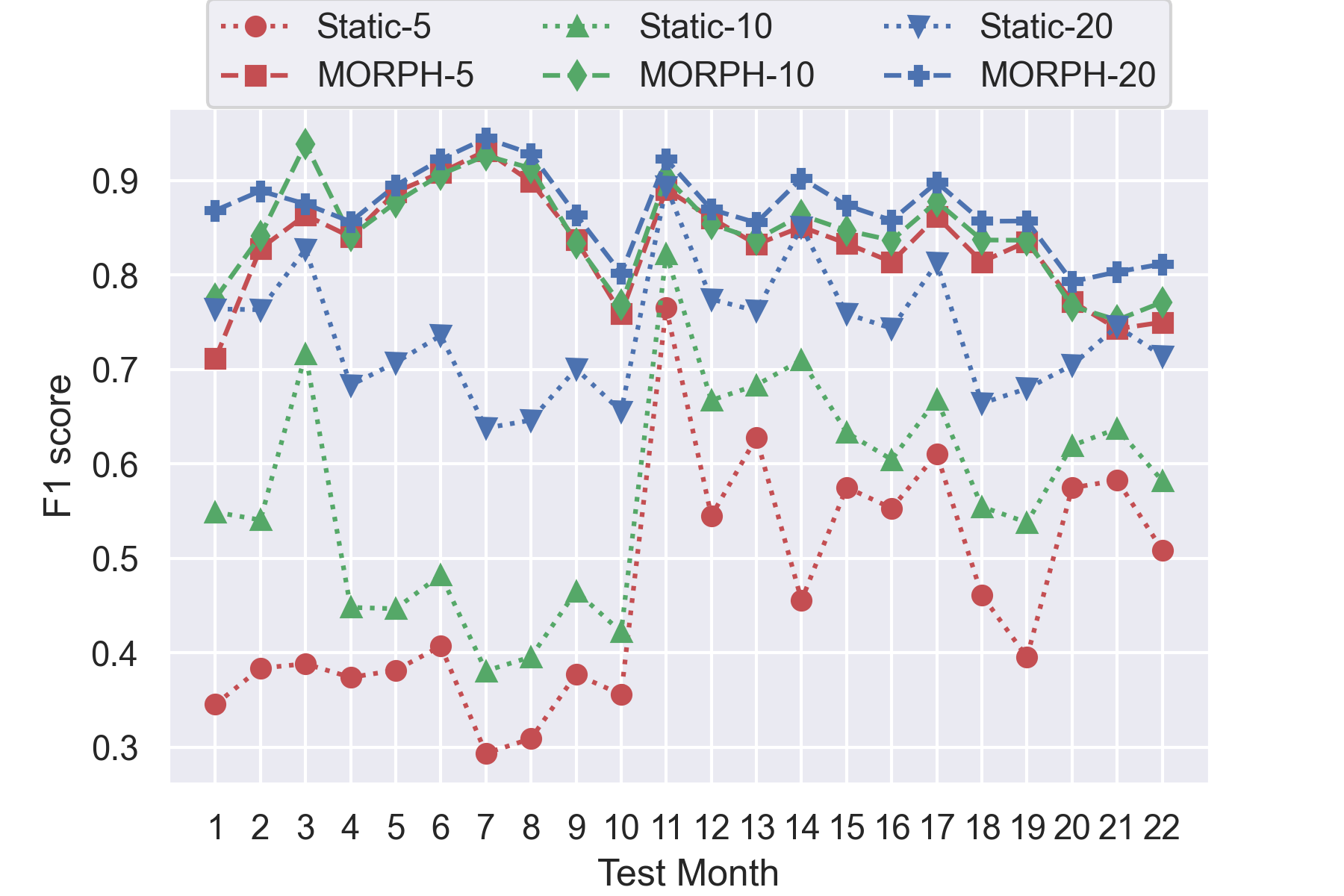}
  \end{subfigure}
  \hfill
  \begin{subfigure}[b]{0.32\textwidth}
    \includegraphics[width=\textwidth]{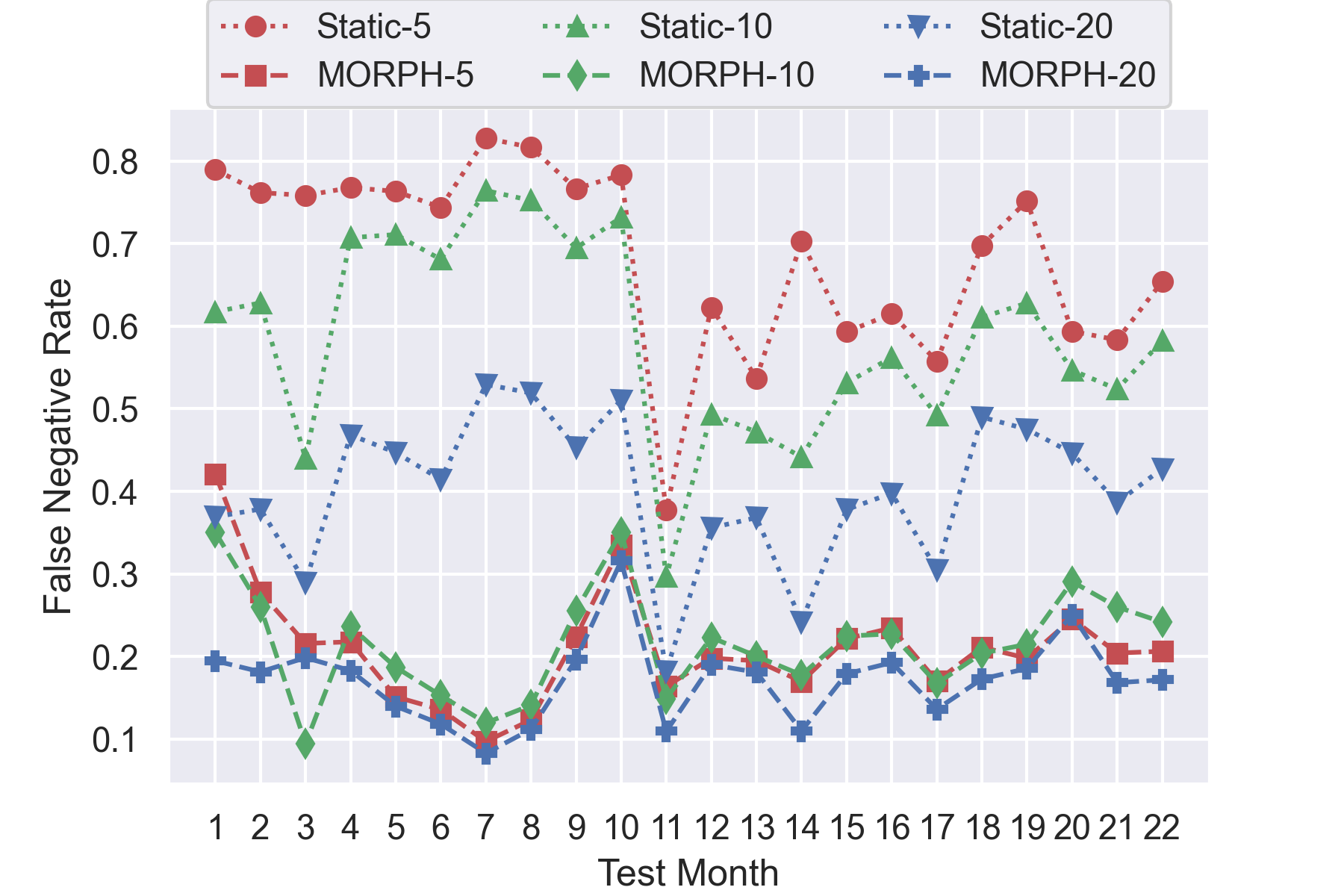}
  \end{subfigure}
    \hfill
  \begin{subfigure}[b]{0.32\textwidth}
    \includegraphics[width=\textwidth]{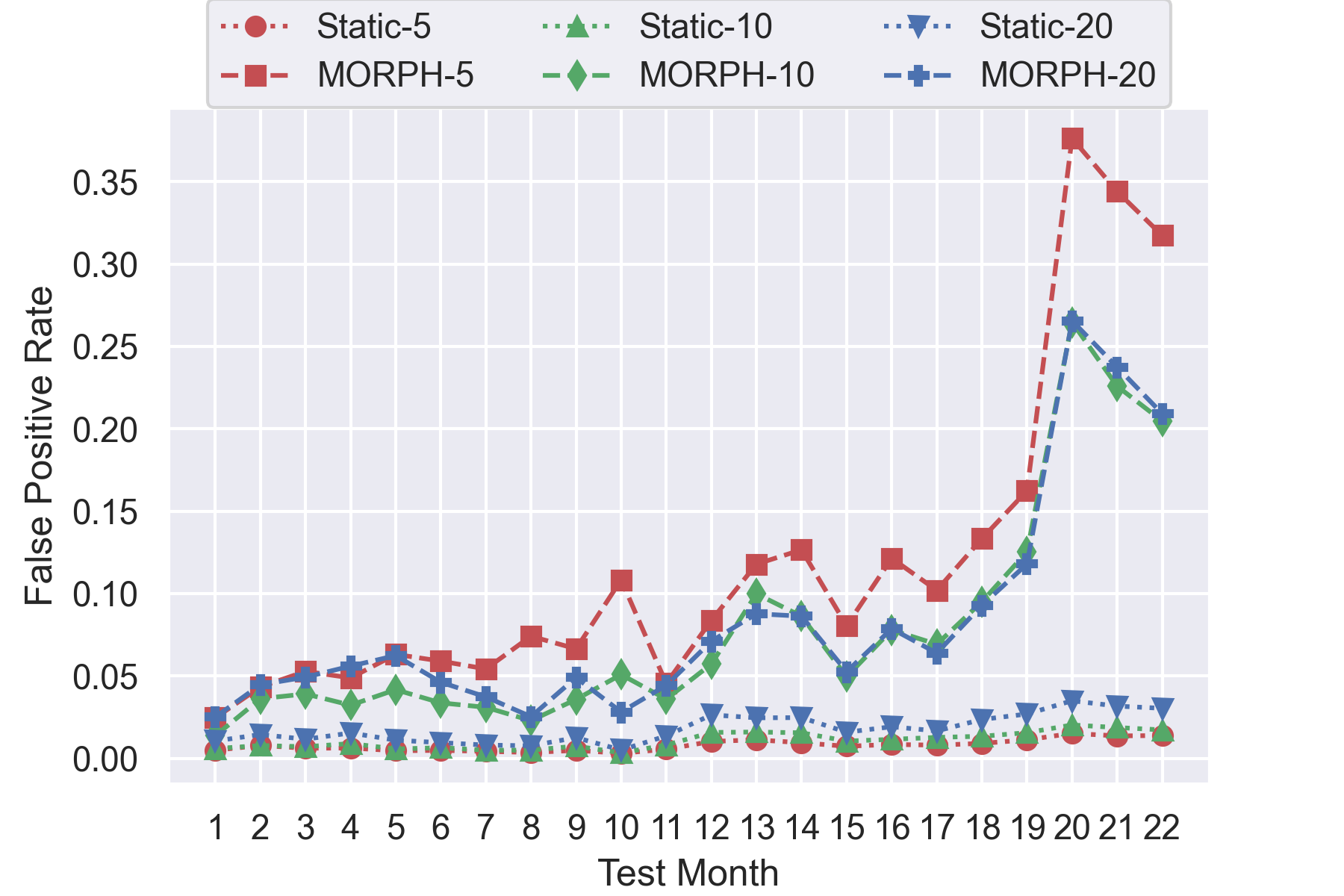}
  \end{subfigure}
  \caption{F1-score (left), FNR (center), and FPR(right) for test months on Ember dataset~\cite{2018arXiv180404637A} with (Adapted) and without (Static) concept drift adaptation method as the number of families in the training dataset is varied. The number in the legend indicates the number of malware families in the training dataset.}
  \label{fig:ember-fam}
\end{figure}

\subsubsection{Impact of Varying Known Malware Families}

To illustrate our second conclusion further, we investigate the impact of varying the number of known malware families in the Ember dataset~\cite{2018arXiv180404637A} on the performance of our concept drift adaptation algorithm. Specifically, we explore the effect of selecting samples exclusively from the top-k most frequent malware families for the training data while retaining all samples in the test data. By limiting the training data to samples from a subset of the most prevalent malware families, we simulate a scenario where the model has limited exposure to diverse malware patterns during training.

We show the results in Figure~\ref{fig:ember-fam}. When the training data contains smaller malware families, the test data exhibits a noticeable concept drift for the malware class. As a result, the model's performance suffers, leading to high false negative rates and reduced overall F1 score. However, our adaptation algorithm effectively mitigates the impact of concept drift. By incorporating the selected samples from the limited set of malware families into the training process, the algorithm adapts the model to the changing malware patterns present in the test data. We observe improvements in the model's performance as it becomes more adept at detecting and classifying malware instances despite the evolving nature of the malware families. We notice a significant improvement in the F1 score (e.g., 36.6\% for five malware families). However, this naturally leads to a larger FPR, especially in the later months compared to the baseline model.

\subsection{Combination with Active Learning}
To address our research question RQ2, we conducted experiments that combined the proposed method with active learning. Specifically, we sampled the most uncertain predictions (i.e., lowest probability output from the softmax layer) for active learning annotations. The number of annotated samples depended on the available human effort for this task. Ideally, we would prefer to simulate a scenario where samples arrive daily, which would closely resemble real-world conditions~\cite{ugarte2019close}. However, since our data is only available at a monthly granularity, we performed active learning at specific month intervals. During the active learning process, we first selected samples from the test data that required annotation and then added them to the training data with their annotated ground truth labels.

\begin{figure}[t]
  \centering
  \begin{subfigure}[b]{0.32\textwidth}
    \includegraphics[width=\textwidth]{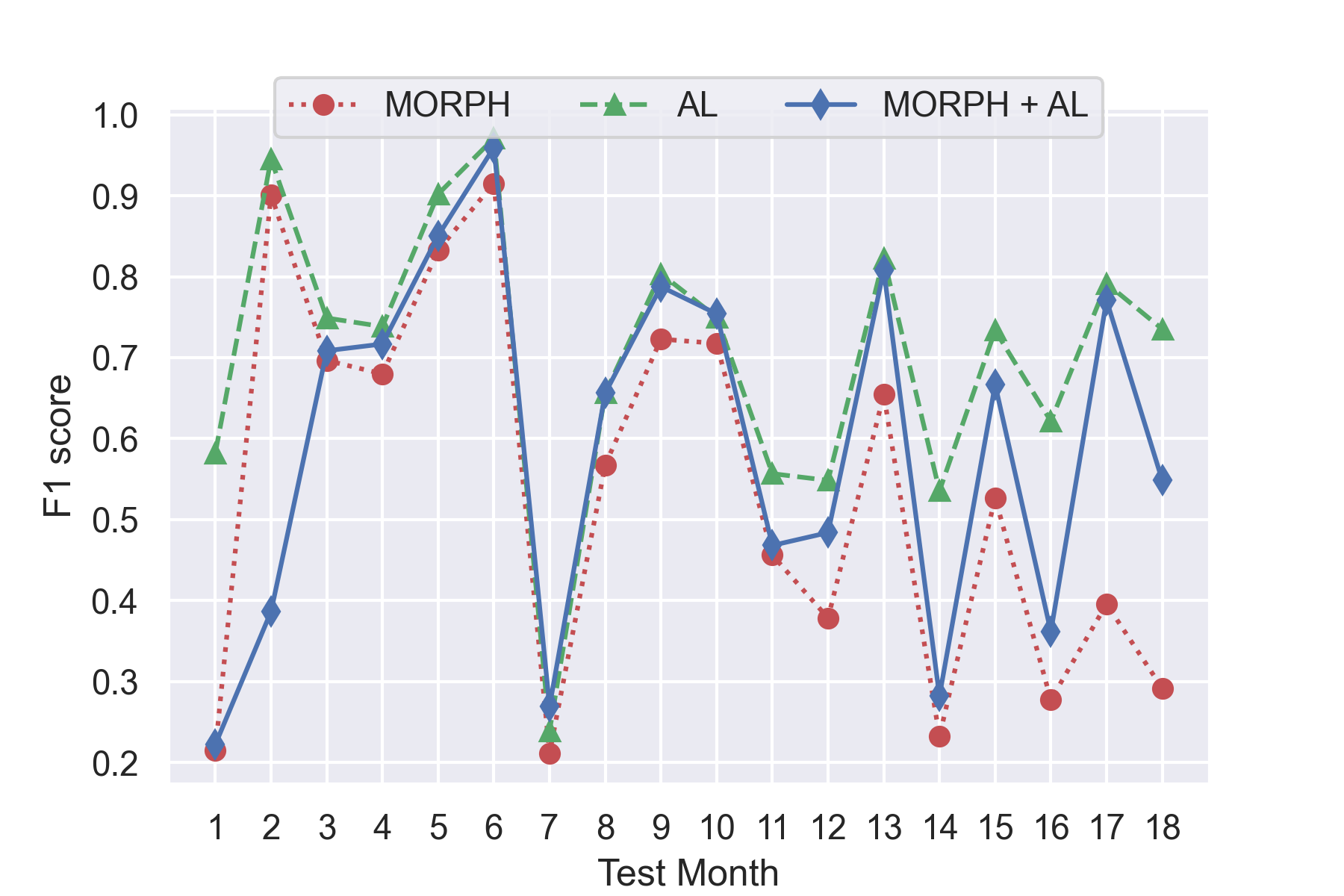}
  \end{subfigure}
  \hfill
  \begin{subfigure}[b]{0.32\textwidth}
    \includegraphics[width=\textwidth]{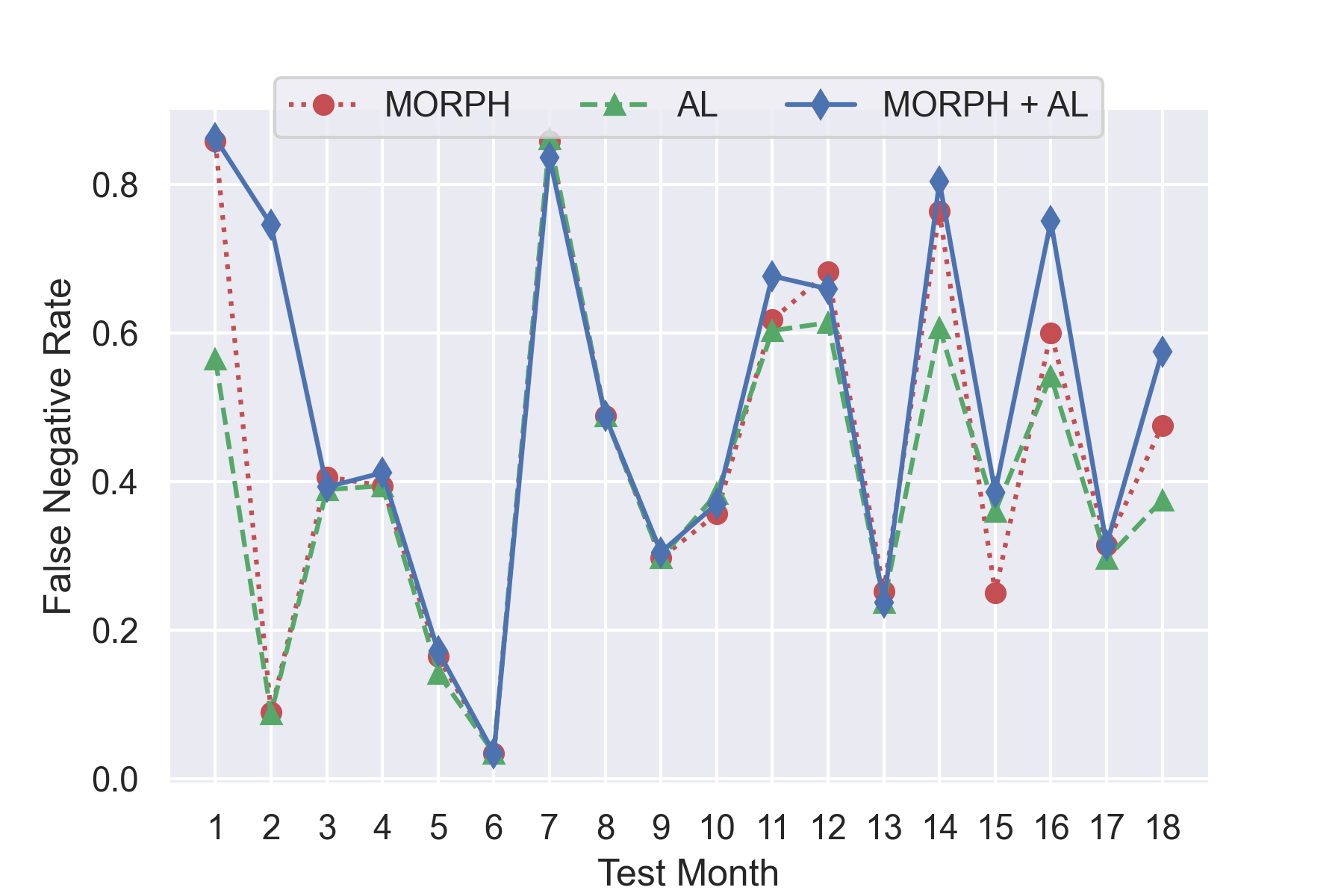}
  \end{subfigure}
    \hfill
  \begin{subfigure}[b]{0.32\textwidth}
    \includegraphics[width=\textwidth]{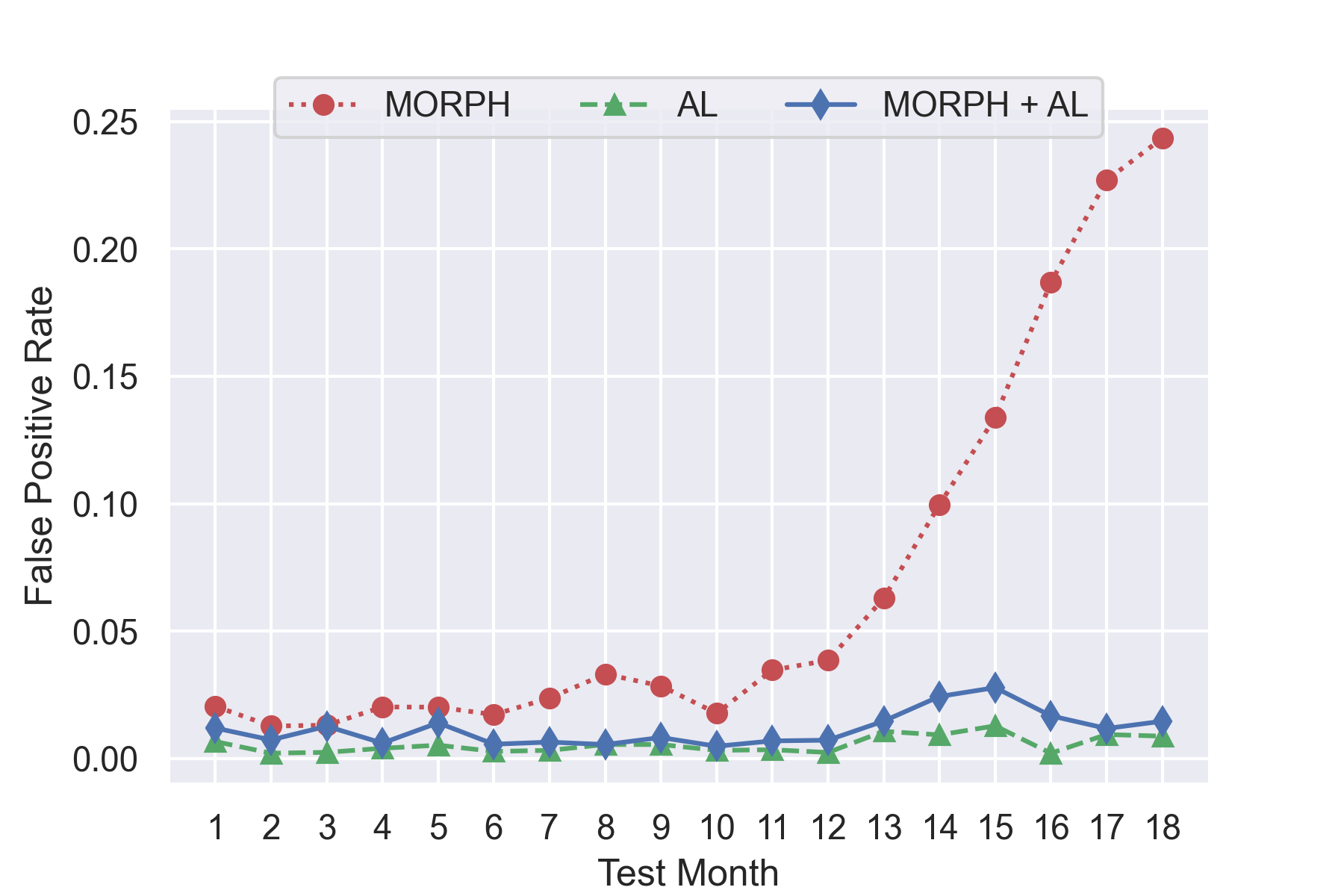}
  \end{subfigure}
  \caption{F1-score (left), FNR (middle), and FPR (right) as active learning (AL) is combined with MORPH. AL: only active learning is applied each month, MORPH+AL: Active learning and MORPH are applied every other month.}
  \label{fig:acl}
\end{figure}


We show the result in \ref{fig:acl} as we reduce the active learning updates and replace them with pseudo-label updates. For active learning updates, we annotate 100 samples per month. We then replace the active learning update with MORPH-based update updates when simulating the effect when both are used in tandem. This effectively reduces the number of active learning updates by 50\%. The result suggests the model can retain comparable performance with regular active learning updates, even when we periodically replace them with pseudo-label-based updates.

\subsubsection{}

\begin{figure}[t]
  \centering
  \begin{subfigure}[b]{0.32\textwidth}
    \includegraphics[width=\textwidth]{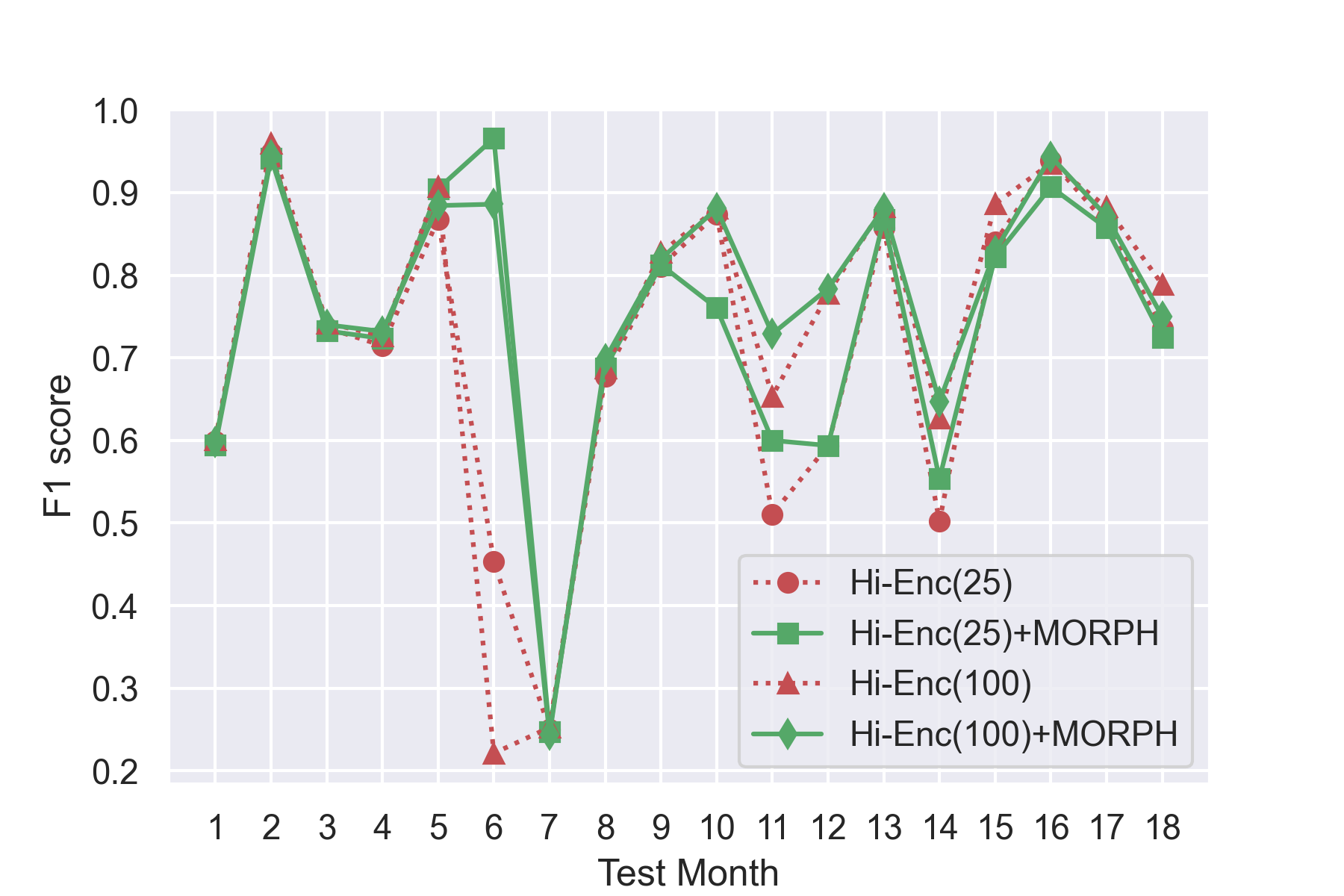}
  \end{subfigure}
  \hfill
  \begin{subfigure}[b]{0.32\textwidth}
    \includegraphics[width=\textwidth]{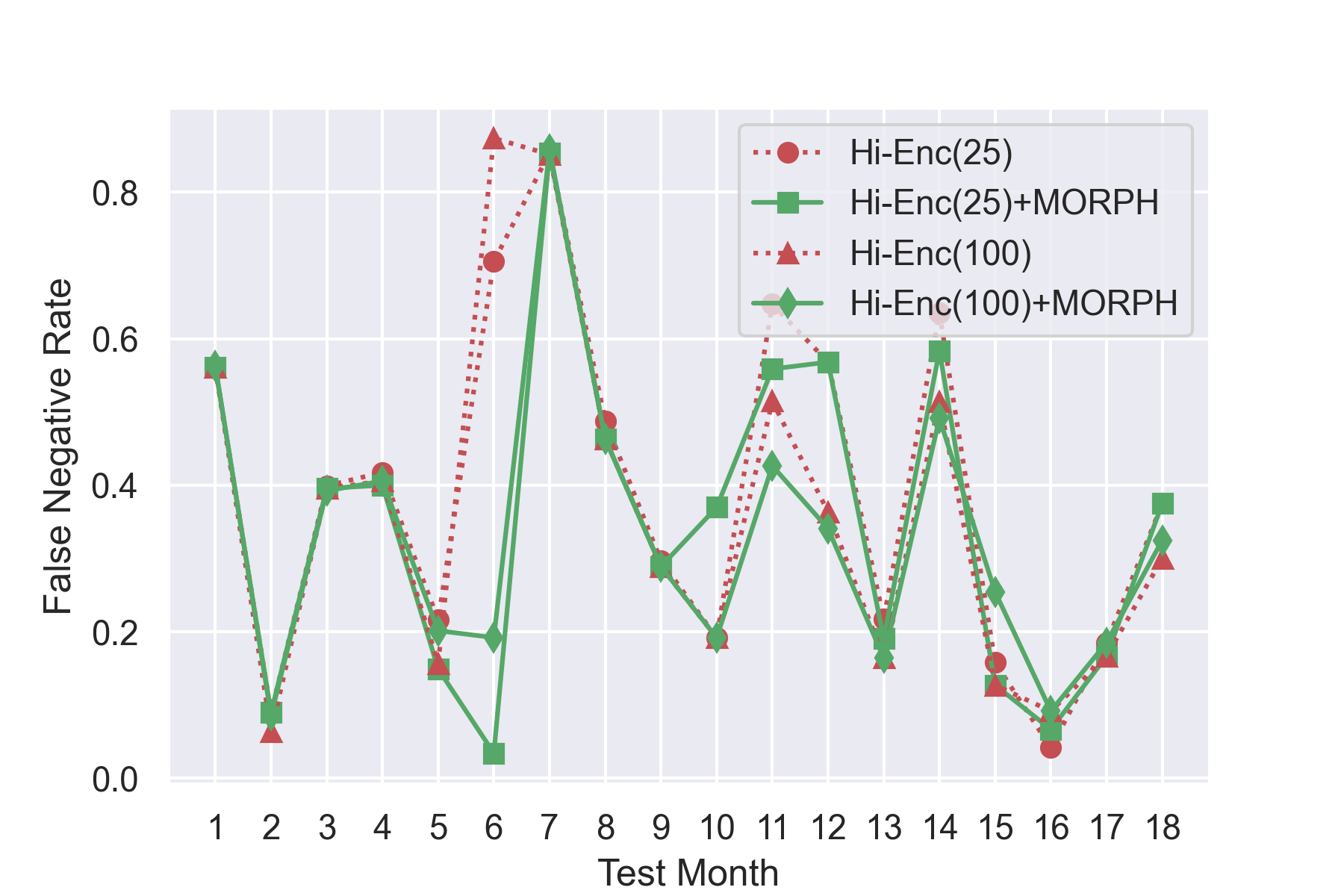}
  \end{subfigure}
    \hfill
  \begin{subfigure}[b]{0.32\textwidth}
    \includegraphics[width=\textwidth]{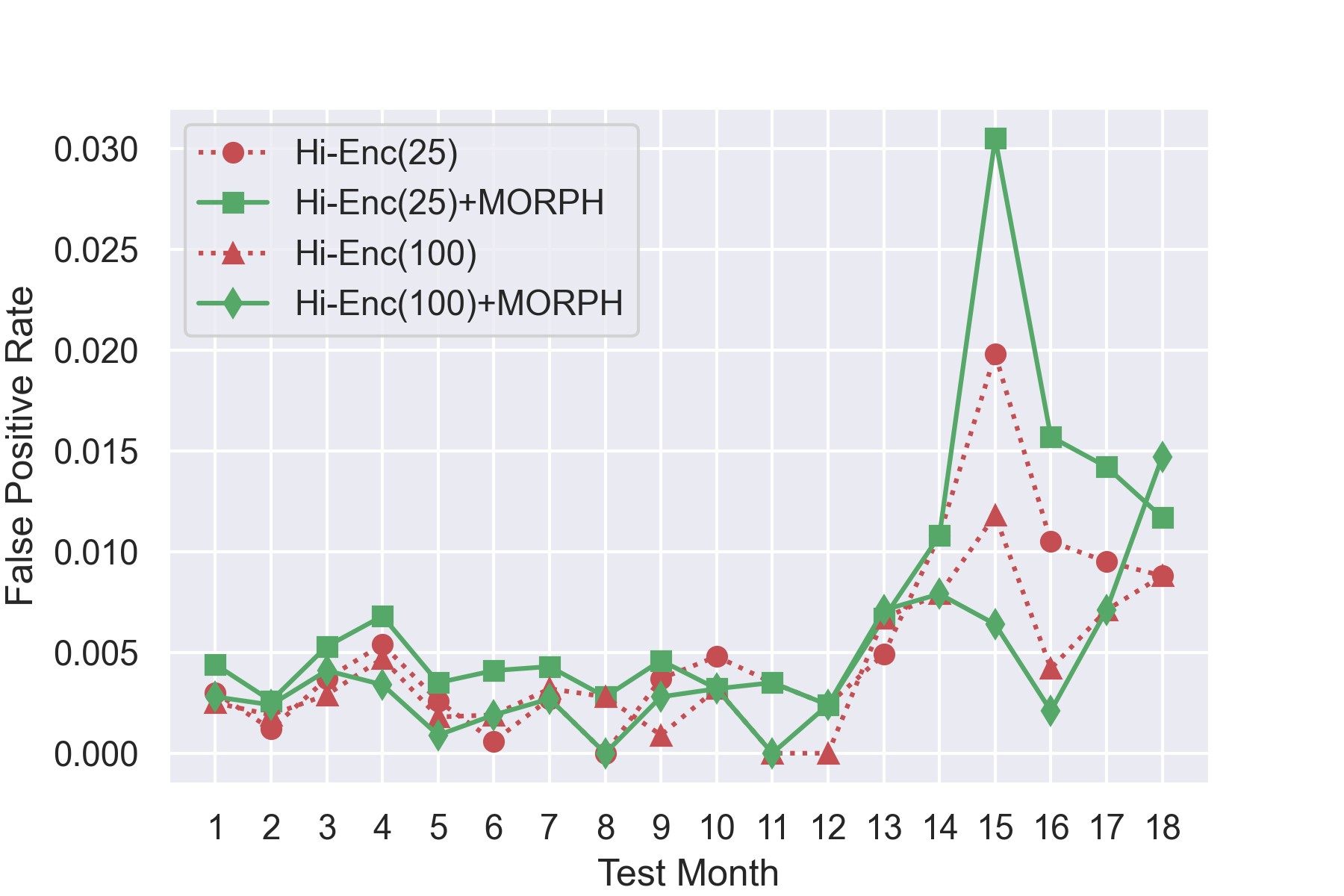}
  \end{subfigure}
  \caption{F1 score (left), FNR (middle), and FPR(right) for Hi-Enc (Hierarchical Contrastive Learning) ~\cite{chen2023continuous} active learning with MORPH updates. The numbers in brackets indicate the amount of labeled samples used for active learning updates each month.}
  \label{fig:hienc}
\end{figure}

\subsubsection{Utility of Unlabeled Data:}
We conduct additional experiments to assess the effectiveness of our proposed algorithm in conjunction with state-of-the-art neural networks and active learning methods. Specifically, we adopt the approach introduced by Chen et al.~\cite{chen2023continuous} and treat the unannotated samples after active learning update as unlabeled data. Subsequently, we employed MORPH to update the model using monthly pseudo-labels assigned to these unlabeled samples. With successive active learning and semi-supervised updates each month, we expect the model to be able to adjust to both sudden and gradual shifts in data distribution. The results, depicted in Figure~\ref{fig:hienc} for the AndroZoo dataset, demonstrate that MORPH consistently enhances performance by leveraging the unlabeled samples instead of only performing active learning updates. When using a 100 annotation budget for active learning, MORPH improves the F1 score by 3.46\%. It is important to consider that the AndroZoo dataset typically consists of 300 or fewer malware samples each month. Thus, we anticipate even more promising outcomes as the number of unlabeled malware samples increases in the dataset.

\textbf{\textit{We arrive at the following conclusions for our RQ2}}
\begin{enumerate}
    \item MORPH can reduce the number of active learning updates required to maintain stable model performance.
    \item Active learning can mitigate the effect of self-poisoning introduced by noisy pseudo labels.
    \item MORPH can be used with other active learning methods to improve model performance.
\end{enumerate}

\subsection{Comparison With DroidEvolver++}
\label{sec:de-cmp}

DroidEvolver~\cite{xu2019droidevolver} and its updated version, DroidEvolver++~\cite{kan2021investigating}, are significant prior works in automated malware concept drift adaptation. These methods operate by maintaining a set of models and adapting to changes in data distribution as the model pool evolves. However, comparing our approach with DroidEvolver is challenging due to the different model architectures used. Nonetheless, we compare our experimental results with the DE model update using pseudo labels.

\begin{figure}[t]
  \centering
  \begin{subfigure}[b]{0.32\textwidth}
    \includegraphics[width=\textwidth]{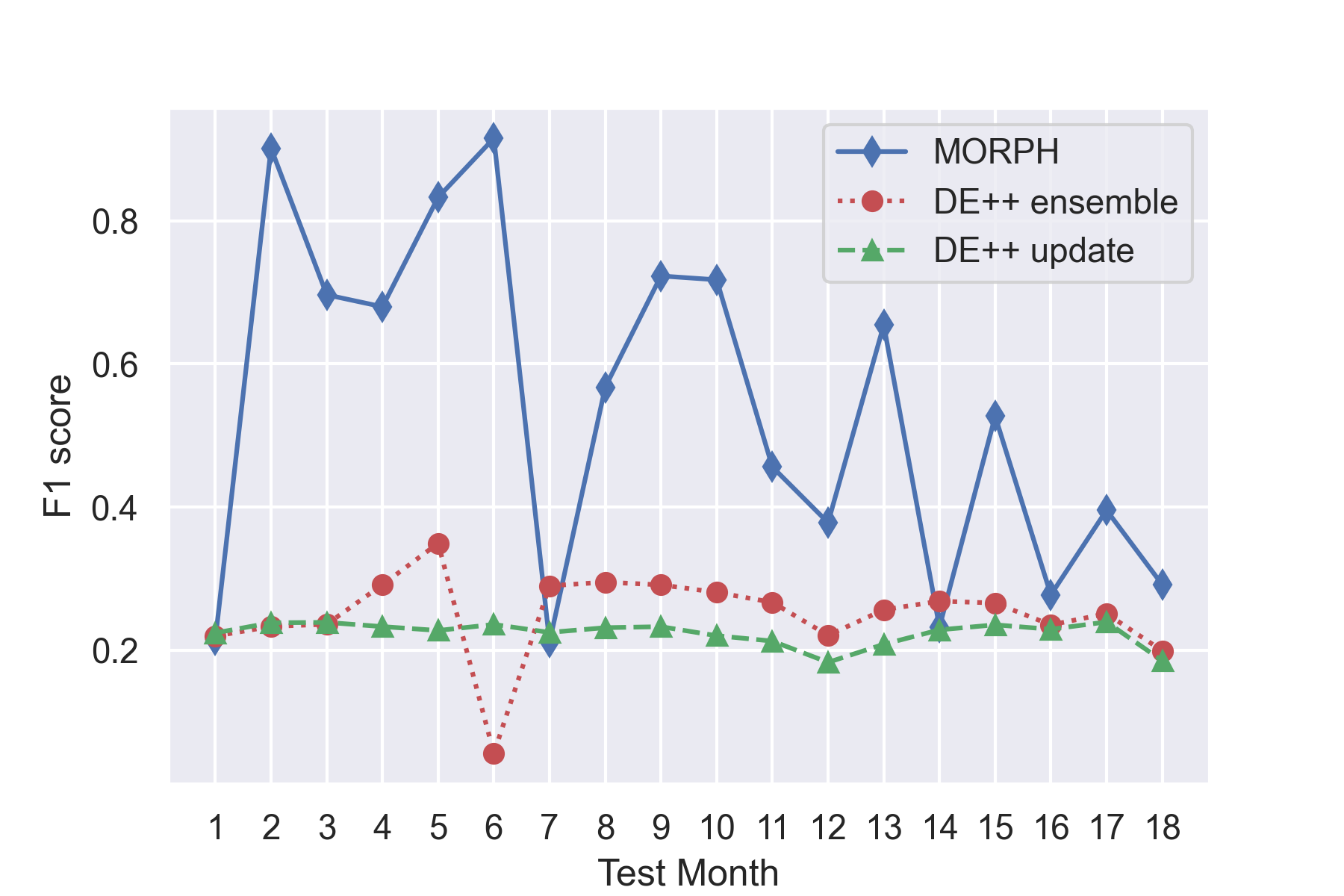}
  \end{subfigure}
  \hfill
  \begin{subfigure}[b]{0.32\textwidth}
    \includegraphics[width=\textwidth]{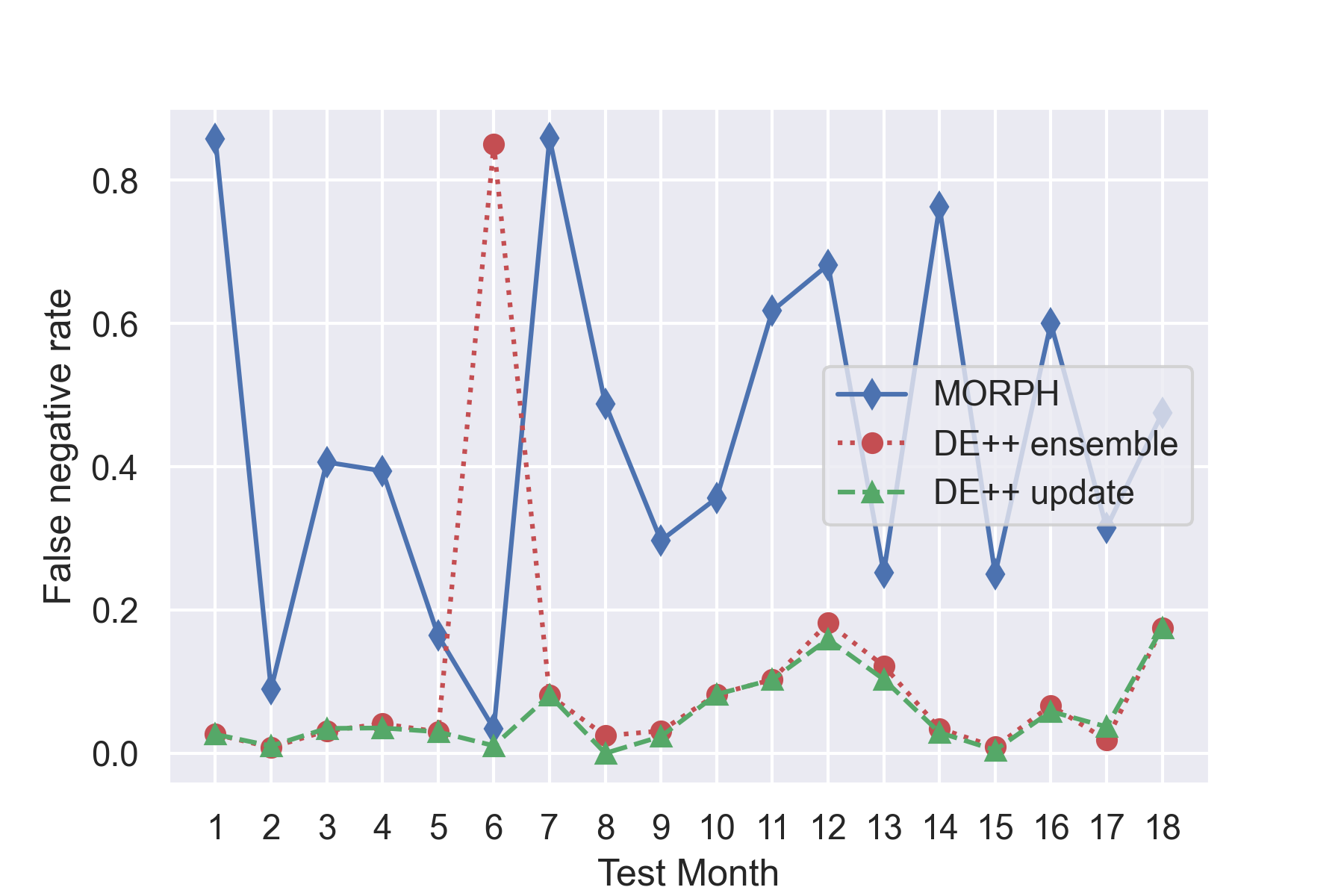}
  \end{subfigure}
    \hfill
  \begin{subfigure}[b]{0.32\textwidth}
    \includegraphics[width=\textwidth]{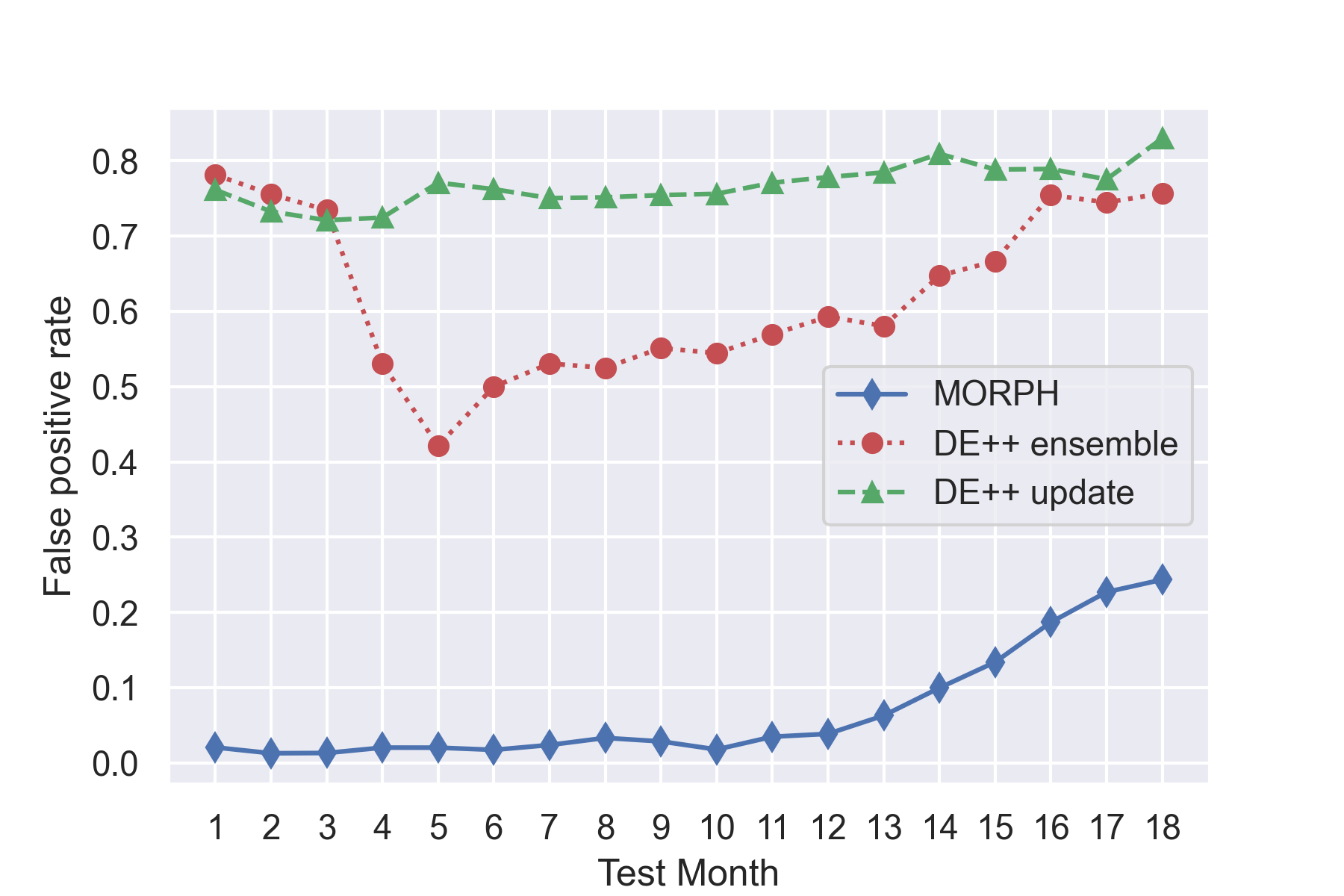}
  \end{subfigure}
  \caption{F1 score (left), FNR (middle), and FPR(right) for DE++ ensemble model with and without model updates, compared against MORPH.}
  \label{fig:de}
\end{figure}

DroidEvolver utilizes an ensemble of five linear online learning models.
The ensemble is trained using an initial dataset of labeled malware and goodware. During testing, DroidEvolver aggregates the predictions of the underlying models using a weighted sum of decision scores as the ensemble decision function. Predictions of individual models that deviate from the ensemble predictions are excluded. The model is then retrained using samples that differ from the ensemble prediction, treating the ensemble prediction as the ground truth annotation. 
Further details about DE++ can be found in the original paper by Kan et al.~\cite{kan2021investigating}. 
Figure~\ref{fig:de} shows the result for DE++ performance, with and without model updates. The linear models do not perform well as concept drift settles in, and the ensemble-based model update further deteriorates their performance by 2.65\% F1 score. On average, DE++ achieves a 23.3\% F1 score, whereas our method achieves a significantly higher F1 score of 53.8\%.

\textbf{\textit{We arrive at the following conclusions regarding our RQ3:}}
\begin{enumerate}
    \item Neural networks are more robust to concept drift than linear online learning models or an ensemble of such models.
    \item Our proposed method is more robust to self-poisoning than a pseudo-label-based algorithm applied to an ensemble of linear classifiers.
\end{enumerate}


%% file: sections/discussions.tex
\section{Discussion}

\subsection{Difference in Datasets}
Our results demonstrate that the impact of concept drift is less severe in the Windows dataset. Even when introducing concept drift by limiting the number of malware families in the training dataset, MORPH can adapt to the new distribution. The primary reason for this is the scale of the dataset. The total number of introduced malware samples in each test month is much smaller in AndroZoo~\cite{Allix:2016:ACM:2901739.2903508} compared to the EMBER dataset~\cite{2018arXiv180404637A}, approximately 100 times smaller. The large number of malware samples in the EMBER dataset~\cite{2018arXiv180404637A} allows for more variations of each new malware family to be included. As a result, the model can correctly predict some of these variations, even when encountering an entirely new family. However, in AndroZoo~\cite{Allix:2016:ACM:2901739.2903508}, the model will likely fail to accurately identify any samples from a new malware family, resulting in their exclusion from the training phase. The number of samples for different families in the training set is significantly lower in AndroZoo~\cite{Allix:2016:ACM:2901739.2903508}, with only 45 of 122 families having at least five samples. This indicates that AndroZoo~\cite{Allix:2016:ACM:2901739.2903508} has a relatively higher number of samples from new families during testing, leading to a higher degree of concept drift, as observed in our experimental results.



\subsection{Feasibility of Complete Automation}
While MORPH shows promising results, fully automated concept drift adaptation is not always possible. When a new malware family exhibits behavior that significantly differs from the samples in the training set, adapting to these samples without manual annotation may not be feasible. This is because we need at least some of these samples to be correctly pseudo-labeled to incorporate them in the training with accurate annotation. It can also violate the gradual drift assumption discussed in Section~\ref{sec-motivation}. Active learning becomes the only viable approach to identify and include these samples in the training data with the appropriate ground truth annotation.



\subsection{Need for Better Features}
Using API usage-based features for malware detection presents a significant drawback: adversaries can evade detection by incorporating spurious API usage into their malicious code.  These shortcomings of malware detection based on static analysis have been studied extensively in prior literature~\cite{moser2007limits,aghakhani2020malware}. One potential avenue for future research in this direction involves leveraging Transformer-based~\cite{vaswani2017attention} models to learn features directly from the source code~\cite{rahali2021malbert,seneviratne2022self}. By focusing on the behavioral aspects rather than relying solely on API usage, detection can become more effective and resistant to concept drifts. 

%% file: sections/conclusion.tex
\section{Conclusion}
This research addresses the critical challenge of concept drift in malware detection, where trained models struggle with evolving threats. We propose a novel self-training approach utilizing pseudo-labels on a neural network-based malware classifier, achieving automatic adaptation and significantly reducing annotation demands compared to traditional active learning. Our method outperforms existing works, effectively handling diverse malware landscapes on both Android and Windows datasets. However, fully automating adaptation might not always be feasible for significantly different malware families. To address this, future research can explore features like behavioral patterns and Transformer-based models for enhanced robustness and effectiveness against evolving threats. Finally, this research paves the way for next-generation malware detection systems capable of continuous adaptation and protection against emerging dangers.